\pdfoutput=1
\documentclass[11pt]{article}

\usepackage[]{acl}
\usepackage{times}
\usepackage{latexsym}
\usepackage{amsmath}
\usepackage{float}
\usepackage{stfloats}
\restylefloat{table}

\usepackage{tcolorbox}
\usepackage{tabularx}
\usepackage{array}
\usepackage{colortbl}
\tcbuselibrary{skins}

\newcolumntype{Y}{>{\raggedleft\arraybackslash}X}
\tcbset{tab1/.style={fonttitle=\bfseries\large,fontupper=\normalsize\sffamily,
colback=yellow!10!white,colframe=red!75!black,colbacktitle=Salmon!40!white,
coltitle=black,center title,freelance,frame code={
\foreach \n in {north east,north west,south east,south west}
{\path [fill=red!75!black] (interior.\n) circle (3mm); };},}}

\tcbset{tab2/.style={enhanced,fonttitle=\bfseries,fontupper=\normalsize\sffamily,
colback=yellow!10!white,colframe=red!50!black,colbacktitle=Salmon!40!white,
coltitle=black,center title}}

\usepackage[T1]{fontenc}
\usepackage[utf8]{inputenc}
\usepackage{multirow}
\usepackage{booktabs}
\usepackage{graphicx}
\usepackage{caption}
\usepackage{subcaption}
\usepackage{amssymb}% http://ctan.org/pkg/amssymb
\usepackage{pifont}% http://ctan.org/pkg/pifont
\newcommand{\cmark}{\ding{51}}%
\newcommand{\xmark}{\ding{55}}%
\usepackage{xcolor}

\usepackage{comment}

\usepackage{microtype}

\usepackage{xcolor}

\DeclareMathOperator*{\argmax}{arg\,max}
\DeclareMathOperator*{\argtopk}{arg\,topk}
\DeclareMathOperator*{\nli}{NLI}
\DeclareMathOperator*{\hvm}{HVM}

\newenvironment{itemizesquish}[2]{\begin{list}{\labelitemi}{\setlength{\itemsep}{#1}\setlength{\labelwidth}{#2}\setlength{\leftmargin}{\labelwidth}\addtolength{\leftmargin}{\labelsep}}}{\end{list}}

\usepackage{hyperref}

\title{Think While You Write:\\Hypothesis Verification Promotes Faithful Knowledge-to-Text Generation}

\author{Yifu Qiu$^1$\thanks{~~Work done while the author was an intern at Apple.}, Varun Embar$^2$, Shay B. Cohen$^1$, Benjamin Han$^2$ \\
  University of Edinburgh$^1$, Apple$^2$ \\
  \{yifu.qiu, scohen\}@ed.ac.uk, \{v\_embar, ben.b.han\}@apple.com
}

\begin{document}
\maketitle
\begin{abstract}

Knowledge-to-text generators often struggle to faithfully generate descriptions for the input facts: they may produce \textit{hallucinations} that contradict the input, or describe facts not present in the input. To reduce hallucinations, we propose a decoding-only method, TWEAK (\textbf{T}hink \textbf{W}hile \textbf{E}ffectively \textbf{A}rticulating \textbf{K}nowledge), which can be integrated with any generator without retraining. TWEAK treats the generated sequences at each decoding step and its future sequences as \textit{hypotheses}, and ranks each generation candidate based on the extent to which their hypotheses are supported by the input facts using a Hypothesis Verification Model (HVM). We first demonstrate the effectiveness of TWEAK by using a Natural Language Inference (NLI) model as the HVM and report improved faithfulness with a minimal impact on the quality. We then replace the NLI model with a task-specific HVM trained with a first-of-a-kind dataset, FATE (\textbf{F}act-\textbf{A}ligned \textbf{T}extual \textbf{E}ntailment), which pairs input facts with their original and perturbed descriptions. We test TWEAK with two generators, and the best TWEAK variants improve on average for the two models by 2.24/7.17 points in faithfulness (FactKB) in in/out-of-distribution evaluations, respectively, and with only a 0.14/0.32-point decline in quality (BERTScore)\footnote{Our code and dataset are at \small{\url{https://github.com/apple/ml-tweak}}.}.
\end{abstract}

% ========================================

\section{Introduction}
\label{sec:intro}

Knowledge-to-text generation (K2T) aims to generate precise and fluent textual descriptions which are consistent with the input facts~\cite{gardent-etal-2017-webnlg, perez-lapata2018, agarwal-etal-2021-TekGen, colas2021eventnarrative}. Although the neural generators are capable of generating fluent and high-quality texts on various tasks~\cite{ribeiro-etal-2021-investigating-LM-G2T-Generation, Zhou2021-dialogGeneration, liu-etal-2022-brio-summ,chen-etal-2022-towards-table2text-generation,qiu-cohen-2022-hiergnn}, one major challenge remains to be \textit{hallucination}~\cite{zhao2020reducing-quantity-hallucinations, maynez-etal-2020-factuality-abs-summ, dziri-etal-2022-faithdial, daheim2023elastic, 10.1162/tacl_a_00563-hallucination-MT}, i.e., the tendency of the models to produce outputs that contradict or are not supported by the inputs. 
% Hallucination greatly hinders the application of neural K2T models because of the potential misinformation produced. 

\begin{figure*}[hbt!]
    \centering
    \includegraphics[width=0.9\linewidth]{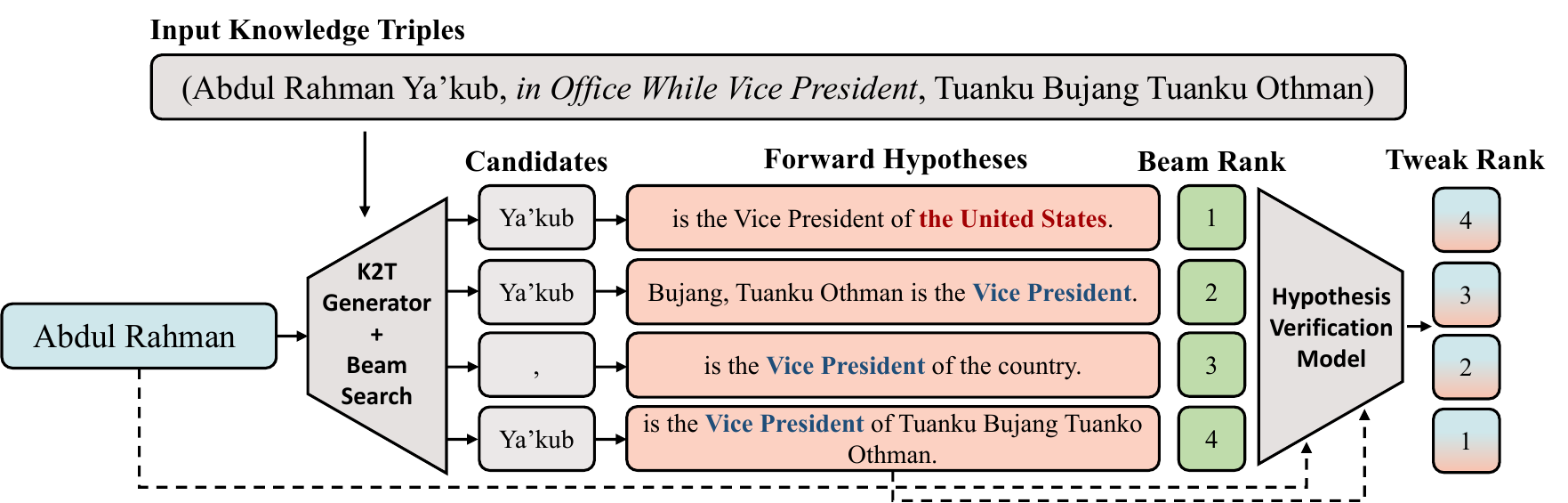}
    \caption{Our proposed TWEAK approach. Compared with beam search which solely ranks the candidates based on generative model's predicted likelihood, TWEAK incorporates \textit{faithfulness}, which is estimated by evaluating the backward and forward hypotheses of each generation candidate with a Hypothesis Verification Model (HVM). In the 4th decoding step of this example, the beam search promotes the candidate leading to hallucinations (e.g., ``United States''), but TWEAK demotes it using signals from HVM.}
    \label{fig:tweak}
\end{figure*}

In this paper, we address the hallucination problem with a \textit{model-agnostic} decoding method, TWEAK (\textbf{T}hink \textbf{W}hile \textbf{E}ffectively \textbf{A}rticulating \textbf{K}nowledge). Different from previous works such as~\cite{hashem-etal-2023-generating}, we \textit{tweak} only the decoding process without requiring re-training of the generative models, thus making our approach easily integratable with any K2T generator. The existing decoding methods of a generative model, such as beam search, sample candidates only from the predicted likelihood without any consideration on the faithfulness implication of these candidates. The problem of \textit{exposure bias} of autoregressive generation only makes the matter worse once any deviation from a faithful generation occurs, since these errors accumulate and become unrecoverable~\cite{schmidt-2019-generalization-exposurebias,zhang2023languageModelSnowball}. TWEAK mitigates this problem by verifying the faithfulness of the candidates at each decoding step to reduce hallucinations. As the example illustrated in Fig.~\ref{fig:tweak}, for each candidate at a decoding step, TWEAK treats the sequence generated so far and its possible future sequence as the \textit{backward} and the \textit{forward hypothesis} (inspired by~\citeauthor{lu-etal-2022-neurologic-astar}), respectively, and feeds them into a Hypothesis Verification Model (HVM) to estimate the candidate's \textit{faithfulness score}, a measure indicating how well the candidate supports the input facts. The candidates are then ranked considering both their generation scores and faithfulness scores.

We first deploy a natural language inference (NLI) model~\cite{nie-etal-2020-adversarial-nli} as the HVM for experimentation, and observe that this approach, TWEAK-NLI, indeed improves the faithfulness of the output compared to the baseline (beam search) by a significant margin. 
% TWEAK-NLI 1) requires a separate consideration for each of the two hypotheses, leading to increased inference cost, and 2) 
The distribution shift between NLI and faithfulness assessment tasks, however, may result in reduced output quality \cite{kryscinski-etal-2020-FactCC,laban-etal-2022-summac,qiu2023detecting-mfact}. We therefore experiment with a second variation, TWEAK-HVM, where we 
propose a task-specific HVM trained with a first-of-a-kind dataset, FATE (\textbf{F}act-\textbf{A}ligned \textbf{T}extual \textbf{E}ntailment). This dataset pairs and aligns input facts with their original and perturbed descriptions. We mimic the autoregressive decoding process where we expand the generation process one token at a time until completion to synthesize the triple-hypothesis pairs with their faithfulness labels. The HVM is then trained to predict all triple-hypothesis labels in a tabular form ~\cite{wang-etal-2021-unire-tableloss, fatahi-bayat-etal-2022-compactie}. Experimental results on WebNLG~\cite{gardent-etal-2017-webnlg} and two out-of-distribution datasets, TekGen~\cite{agarwal-etal-2021-TekGen} and GenWiki~\cite{perez-lapata2018}, confirm the advantages of TWEAK-HVM. It also greatly reduces computation as it encodes both input facts and hypotheses simultaneously.

We summarize our contributions as follows,

\begin{itemizesquish}{-0.3em}{0.5em}
    \item We propose a model-agnostic decoding strategy, TWEAK, which incorporates an HVM for candidate ranking, and show that the approach improves faithfulness of K2T generation when using an NLI model as the HVM.
    \item We propose a new dataset, FATE, which pairs and aligns input facts with their original and perturbed descriptions at word level.
    \item We train a task-specific HVM with FATE and demonstrate its advantages over the NLI-based method
    % , not only 
    in output faithfulness and quality. 
    % but also in the improved inference speed.
    % \item Overall the best TWEAK variants improve on average 2.22/7.17 points on faithfulness measured by FactKB over in/out-of-distribution evaluations, respectively, with only 0.14/0.32 points decrease on quality measured by BERTScore.
\end{itemizesquish}

% ========================================

\section{Related Work}
\label{sec:related_work}

Knowledge-to-text generation tasks involve the transformation of structured data or knowledge into natural language texts \cite{gardent-etal-2017-webnlg, perez-lapata2018, colas2021eventnarrative}. 
Previous works encode the structured input explicitly as models' representations \cite{schmitt-etal-2021-modeling, marcheggiani-perez-beltrachini-2018-deep, guo-etal-2019-densely, rebuffel2020hierarchical, koncel-kedziorski-etal-2019-text}. 
A usual way is to serialize the structured input first and use a pre-trained model to directly generate its description \cite{ribeiro-etal-2021-investigating-plm-for-graph-to-text-generation,li-etal-2021-shot-knowledge,su-etal-2021-shot-table}. However, a notable challenge is hallucinations -- models produce claims that are not supported by inputs \cite{hashem-etal-2023-generating, wang-etal-2023-faithful, yang-etal-2022-tableformer}. Previous work has explored methods including plan-before-generate pipelines \cite{10.1162/tacl_a_00381-marco-plan-generation, 10.1609/aaai.v33i01.33016908, 10.1162/tacl_a_00484}, architecting models to be explicitly fact-aware \cite{wang-etal-2022-robust,ji-etal-2023-rho}, and augmenting the training data with self-supervised learning \cite{han-shareghi-2022-self, wang-etal-2023-faithful, hashem-etal-2023-generating}. 
Mitigating hallucinations in decoding, however, has received relatively less attention, despite its advantages in model-agnostic applications \cite{xiao-wang-2021-hallucination, lu-etal-2022-neurologic-astar, wan-etal-2023-faithfulness}.

Comparing to a recent work~\cite{wan-etal-2023-faithfulness}, where effect of different decoding strategies on faithfulness of \textit{abstractive summarization} is investigated, and a faithfulness re-ranking method is proposed to improve output, our work is unique in that we target a different task (K2T), use hypothesis verification instead of a faithfulness composite metric to guide the ranking, and train a task-specific HVM based on our novel dataset to bring improvement to both faithfulness and quality.

% ========================================

\section{Knowledge-to-Text: Task Definition}
\label{sec:task_defn}

K2T task concerns generating a natural language description $\textbf{y}$ for a list of input facts $\textbf{x}=\langle \ldots, x_i, \ldots \rangle$, where $x_i$ is a fact triple represented as \texttt{<subj, rel, obj>}
% $(\text{subj}, \text{rel}, \text{obj})$
indicating a relation, $\texttt{rel}$, holds between the subject entity, $\texttt{subj}$, and the object entity, $\texttt{obj}$. Two complementary requirements exist for an ideal generation: a \textit{high-quality} generation should describe all of the input facts in a grammatical and readable fashion, while a \textit{faithful} generation should not add any additional claim or contradict any input fact. 
% For example, given the fact triples, {\small \texttt{(Joe\_Biden, birth\_place, Scranton,\_PA)}} and {\small\texttt{(Joe\_Biden, date\_of\_birth, 1942-11-20)}}, ``\textit{Joe Biden was born in Scranton, PA on November 20, 1942}'' is a high-quality and faithful description.

We use an autoregressive language model $p_{\theta}$ with parameters $\theta$ to estimate the probability of the token sequence $\textbf{y}=\langle \ldots, y_t, \ldots \rangle$, i.e., $p_{\theta}(\textbf{y} \mid \textbf{x}) = \prod^{|\textbf{y}|}_{t=1}p_{\theta}(y_t \mid \textbf{y}_{<t}, \textbf{x})$. To decide on the final output, a \textit{decoding} process finds the optimal sequence by solving $\textbf{y}^*=\argmax_{\textbf{y}\in Y}F(\textbf{y})$, where $Y$ is the set of all possible sequences, and $F$ is an objective function. This can be accomplished by selecting the top $k$ candidates generated from vocabulary $\mathcal{V}$ using an $F$-approximating scoring function $f$ one token $y_t$ at a time:
%and extending the partial sequence by concatenating it with each candidate token:
\begin{subequations}\label{equ:argtopk_scoring}
\begin{align}
    Y'_t &= \{\textbf{y}_{<t}\circ y_t\mid \textbf{y}_{<t}\in Y_{t-1}, y_t\in \mathcal{V}\},\nonumber\\
    Y_t &= \argtopk_{(\textbf{y}_{<t}\circ y_t)\in Y'_t} \{f(\textbf{y}_{<t}, y_t, \textbf{x})\}.\tag{\ref{equ:argtopk_scoring}}
\end{align}
\end{subequations}
Common decoding strategies, such as greedy and beam search, set $f$ to $\log p_{\theta}(\textbf{y}_{\leq t} \mid \textbf{x})$. In greedy search $k$  is set to $1$. In Sec.~\ref{subsec:constrained_decoding_HV} we describe our scoring function that promotes faithful generation via hypothesis verification. 

% ========================================

\section{TWEAK}
\label{sec:tweak}

We now describe our approach in Sec.~\ref{subsec:constrained_decoding_HV}, the FATE dataset in Sec.~\ref{subsec:fate}, and our task-specific HVM trained with the dataset in Sec.~\ref{subsec:hvm}.

% ------

\subsection{Decoding with Hypothesis Verification}
\label{subsec:constrained_decoding_HV}

TWEAK is a model-agnostic decoding method that incorporates faithfulness objective into the decoding process. 
As shown in Fig.~\ref{fig:tweak}, at each decoding step we rank a candidate not only by its predicted likelihood from the generator, i.e., $\log p_{\theta}(\textbf{y}_{\leq t} \mid \textbf{x})$, but also by its \textit{faithfulness score}. To assess the faithfulness for a single candidate, we ask the model to look ahead and generate the future sequence until the end \cite{lu-etal-2022-neurologic-astar}, and we approximate the candidate's faithfulness based on the sequence generated to the current step, the \textit{backward hypothesis}, and the future sequence, the \textit{forward hypothesis}, using a HVM.

More specifically, we instantiate the scoring function $f(\textbf{y}_{<t}, y_t, \textbf{x})$ in Equ.~\eqref{equ:argtopk_scoring} as follows\footnote{We omit function arguments as `$\cdot$' if context is clear.}:
\begin{align}\label{equ:tweak_scoring}
\begin{split}
    f(\cdot) &= \log p_{\theta}(\textbf{y}_{\leq t} \mid \textbf{x}) + \alpha\cdot f_{\text{faith}}(\cdot),\\
    f_{\text{faith}}(\cdot) &= w_t\cdot h(\textbf{x}, \textbf{y}_{\le t}) + (1 - w_t) h(\textbf{x}, \textbf{y}_{\text{f}}).
\end{split}
\end{align}
The overall score $f$ is thus a weighted sum of the generator's predicted likelihood and faithfulness $f_{\text{faith}}$. The latter, weighted by $\alpha$,\footnote{The weight ${\alpha}$ can be determined on a validation set such that a desirable balance between output quality and faithfulness is achieved. See Fig.~\ref{fig:weighting_effects} for example.} scores how likely a backward and forward hypothesis, $\textbf{y}_{\le t}$ and $\textbf{y}_{\text{f}}$ respectively, supports the input facts via the hypothesis scoring function $h$, and returns a weighted sum of the faithfulness scores of the two hypotheses. Depending on the implementation of $h$, we have different instantiations for $\textbf{y}_{\text{f}}$ and weight $w_t$, as described in Sec.~\ref{subsubsec:hv_nli} and Sec.~\ref{subsubsec:hv_hvm}.

% ------

\subsubsection{Hypothesis Verification via NLI}
\label{subsubsec:hv_nli}

One simple way to implement an HVM is to treat the concatenated input facts as a premise and the (possibly partial) generated sequence as the hypothesis, then use an NLI model's prediction as the faithfulness score. We thus instantiate Equ.~\eqref{equ:tweak_scoring} as:
\vspace{-0.5cm}

\begin{align}\label{equ:hv_nli}
\begin{split}
    h(\textbf{x}, \textbf{y}) &= \nli(x_1\circ\ldots\circ x_m, \textbf{y}),\\
    \textbf{y}_{\text{f}} &= \textbf{y}_{\le t}\circ g(\textbf{y}_{\le t}, \textbf{x}),\\
    g(\textbf{y}_{\le t}, \textbf{x}) &= \argmax_{\textbf{y}\in\{\textbf{y}_{>t}\}}(\prod^{|\textbf{y}|}_{t'=t+1}p_{\theta}(y_t'|\textbf{y}_{<t'}, \textbf{x})),\\
    w_t &= 
    \begin{cases}
    1 & \small{\text{for TWEAK-NLI-B}} \\
    0 & \small{\text{for TWEAK-NLI-F}} \\
    \frac{t}{|\textbf{y}_{\text{f}}|} & \small{\text{for TWEAK-NLI-B+F.}}
    \end{cases}
\end{split}
\end{align}
The hypothesis scoring function in the above is simply an NLI model returning a score indicating how likely the hypothesis is supported by the premise.\footnote{We only use the the \texttt{entailment} score and discard the scores of \texttt{neutral} and \texttt{contradiction}.} The forward hypothesis $\textbf{y}_{\text{f}}$ is a \textit{complete} sequence concatenating the sequence generated so far and a possible future sequence. Function $g$ is a \textit{greedy} generator producing a future sequence from time step $(t + 1)$ on. We experiment with three NLI-based variants: TWEAK-NLI-B uses only the backward hypothesis with $w_t$ set to 1, TWEAK-NLI-F uses only the forward hypothesis with $w_t$ set to 0, and TWEAK-NLI-B+F uses both, with $w_t$ dynamically set to the ratio of the lengths of the backward and the forward hypotheses at time step $t$. We call this last weighing scheme \textit{dynamic aggregation} (DA), and the intuition is to place less weight on the relatively incomplete backward hypothesis at the early stage of decoding.

% ------

\subsubsection{Hypothesis Verification via HVM}
\label{subsubsec:hv_hvm}

Alternatively, we train a task-specific HVM to score hypotheses, and instantiate Equ.~\eqref{equ:tweak_scoring} as:
\vspace{-0.1cm}
\begin{align}\label{equ:hv_hvm}
\begin{split}
    h(\textbf{x}, \textbf{y}) &= \hvm(\textbf{x}, \textbf{y}),\\
    \textbf{y}_{\text{f}} &= g(\textbf{y}_{\le t}, \textbf{x}),\\
    w_t &= \frac{t}{t+|\textbf{y}_{\text{f}}|}.
\end{split}
\end{align}
Comparing to the NLI-based hypothesis scoring function in Equ.~\eqref{equ:hv_nli}, here we use HVM to compute a score indicating how well sequence $\textbf{y}$ supports input facts $\textbf{x}$. We also consider \textit{only} the future sequence as $\textbf{y}_{\text{f}}$, and the weight $w_t$ is computed entirely dynamically, similar to TWEAK-NLI-B+F. More details of HVM are discussed in Sec.~\ref{subsec:hvm}.

% ------

\subsection{Fact-Aligned Textual Entailment Dataset}
\label{subsec:fate}

To train the task-specific HVM (see Sec.~\ref{subsec:hvm}), we construct a novel dataset called FATE, where each instance is a tuple \texttt{(F$^{+}$, F$^{-}$, T$^{+}$, T$^{-}$)}: \texttt{F$^{+}$, F$^{-}$} are fact triples and their \textit{perturbed} version, and \texttt{T$^{+}$, T$^{-}$} are their respective descriptions. We take \texttt{F$^{+}$} and \texttt{T$^{+}$} from WebNLG~\cite{gardent-etal-2017-webnlg}, and employ a large language model (LLM) \footnote{We use \textit{text-davinci-003}. The prompt templates we use for manipulating triple and description are in Appendix \ref{appendix:prompt-template}.} to perturb one triple in \texttt{F$^{+}$} to construct \texttt{F$^{-}$}. The perturbation may happen in any position in a fact triple, i.e., subject, object, or relation. We then ask the LLM to generate description \texttt{T$^{-}$} for \texttt{F$^{-}$} that is as close to \texttt{T$^{+}$} as possible. The perturbed span is then identified and clearly marked with tag ``\texttt{<S\textit{i}>}'' in both \texttt{T$^{+}$} and \texttt{T$^{-}$}, where \texttt{\textit{i}} indicates the perturbed triple corresponding to the span. We present an instance in Table~\ref{tab:fate_hypos_examples} and the dataset statistics in Appendix~\ref{appendix:fate-stats}.

% \begin{align}\label{equ:fate_example}
% \begin{split}
%     \texttt{F$^{+}$} &= \small{\texttt{(Ireland, {\color{blue}largest\_city}, Dublin)}}\\
%     \texttt{F$^{-}$} &= \small{\texttt{(Ireland, {\color{red}national\_capital}, Dublin)}}\\
%     \texttt{T$^{+}$} &= \small{\textit{"Dublin is Ireland's <S0> {\color{blue}largest city} </S0>"}}\\
%     \texttt{T$^{-}$} &= \small{\textit{"Dublin is Ireland's <S0> {\color{red}national capital}  </S0>"}}\\
% \end{split}
% \end{align}
% Table~\ref{tab:fate} describes the basic stats of the FATE dataset. We describe how FATE is used to train a task-specific HVM in Sec.~\ref{subsec:hvm}.

% ------

\subsection{A Task-specific HVM}
\label{subsec:hvm}

\begin{figure}[t]
    \centering
    \includegraphics[width=0.8\linewidth]{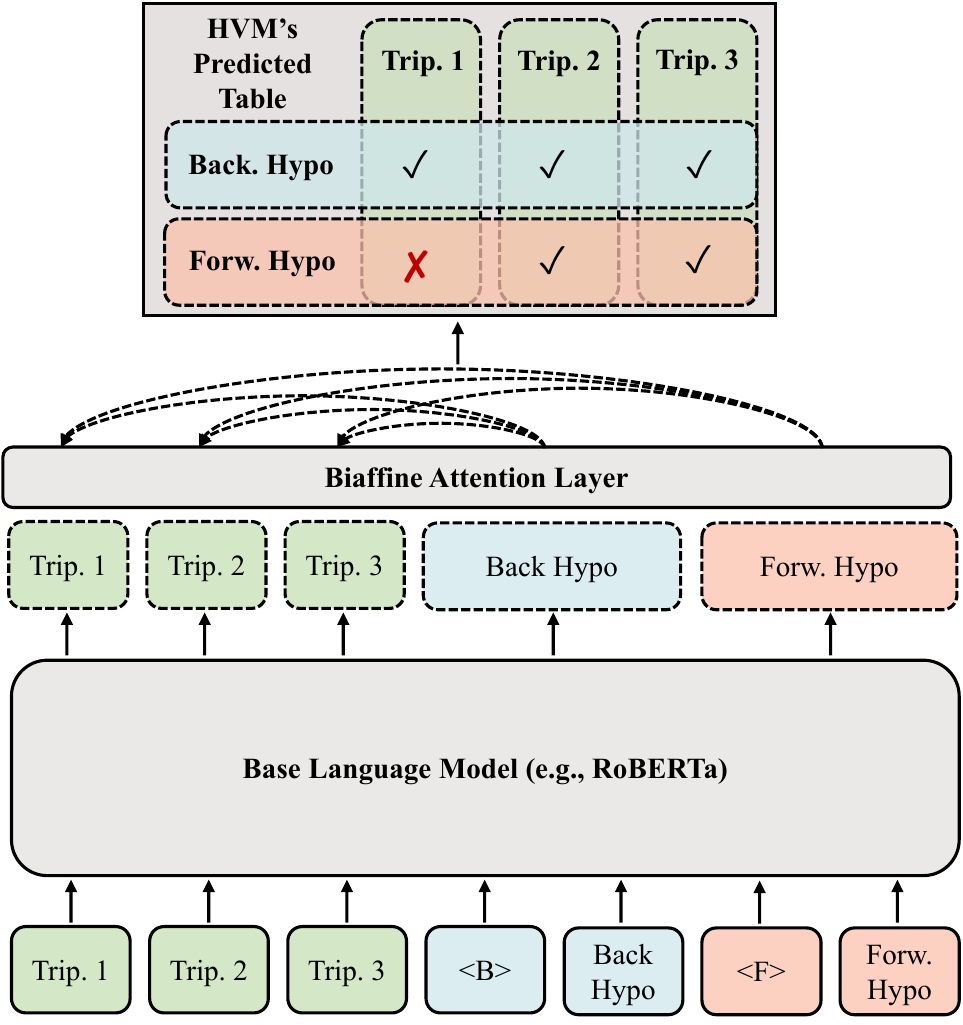}
    \caption{Our task-specific hypothesis verification model. It takes fact triples and backward/forward hypotheses as input, and predicts pair-wise faithfulness relations for each triple-hypothesis pair in a 2D table.}
    \label{fig:hvm}
\end{figure}

There are two disadvantages when using an NLI model as the HVM in TWEAK: 1) the NLI model concatenates all triples into a single premise, losing the entailment relationship between each individual triple and a hypothesis, and 2) NLI models often perform poorly in faithfulness classification due to their inability to generalize to a different target task~\cite{utama-etal-2022-falsesum,kryscinski-etal-2020-FactCC}.

To address these problems, we train a task-specific HVM using our dataset FATE described in Sec.~\ref{subsec:fate}. 
As depicted in Fig.~\ref{fig:hvm}, we first assemble fact triples and the corresponding pair of backward and forward hypotheses as input. We then encode the input via a language model (RoBERTa; \citealt{liu2019roberta}) and use average pooling over all tokens to obtain the representations of each triple and hypothesis. A biaffine attention layer is then used to predict a 2D table representing the pair-wise faithfulness relations (\texttt{unsupported/supported}) between each triple-hypothesis pair. 
% By encoding all this information in the input, and jointly predicting all triple-hypothesis relations, we reduced the computation cost significantly compared to using NLI for hypothesis verification.
Our model is then trained to minimize a table-form objective  \cite{wang-etal-2021-unire-tableloss,fatahi-bayat-etal-2022-compactie},

{\small
\begin{equation}
    L= - \frac{1}{2|\textbf{x}|} \sum_{x\in \textbf{x}} \sum_{\textbf{y}\in \{\textbf{y}_{\le t}, \textbf{y}_{\text{f}}\}} \log P(\hat{B}_{x,\textbf{y}} = B_{x,\textbf{y}} \mid x, \textbf{y}), \nonumber
\end{equation}
}
where $\textbf{x}$ is the set of fact triples in an instance, $\textbf{y}_{\le t}$ and $\textbf{y}_{\text{f}}$ are a corresponding backward and forward hypotheses, and $B_{x, \textbf{y}}$ and $\hat{B}_{x, \textbf{y}}$ are the ground-truth label and the biaffine model prediction for the triple-hypothesis pair, respectively. For inference, we instantiate the function $\hvm$ in Equ.~\eqref{equ:hv_hvm} as:\footnote{\textbf{1} is the \texttt{supported} label.}
\begin{equation}
\small{\hvm(\textbf{x}, \textbf{y}) = \frac{1}{|\textbf{x}|}\sum_{x\in \textbf{x}}\log P(\hat{B}_{x,\textbf{y}} =\textbf{1} \mid x, \textbf{y})}. \nonumber
\end{equation}

\begin{table}[tb!]
{\footnotesize
\begin{center}
\resizebox{\linewidth}{!}{
\begin{tabular}{@{}lcc@{}}
\toprule
\textbf{FATE Instance} & \textbf{Type}              & \textbf{Label} \\ \midrule

\textit{(Ireland, {\color{blue}largest\_city}, Dublin)}  & PTs   & - \\ 
\textit{(Ireland, {\color{red}national\_capital}, Dublin)}  & NTs   & - \\ 
\textit{Dublin is Ireland's <S0> {\color{blue}largest city} </S0>}  & PD   & - \\ 
\textit{Dublin is Ireland's <S0> {\color{red}national capital} </S0>}  & ND   & - \\  \midrule
\textbf{Synthesized Hypotheses (at 10$^{th}$ Decoding Step)} &               &  \\ \midrule
\textit{Dublin is Ireland's {\color{blue}largest}}       & BH         & \cmark            \\
\textit{{\color{blue}largest city}.}                     & FH          & \cmark            \\
\textit{Dublin is Ireland's {\color{red}national}}      & BH         & \xmark            \\
\textit{{\color{red}national capital}.}                 & FH          & \xmark         \\ \bottomrule
\end{tabular}}
\end{center}
}
\caption{FATE's example (upper panel) and the synthesized hyptheses derived from it (bottom panel). {PTs/NTs} stand for the {\color{blue}positive}/{\color{red}negative} triples. {PDs/NDs} are the {\color{blue}positive}/{\color{red}negative} descriptions, and {BH/FH} are the backward/forward hypothses.
\cmark and \xmark indicate {\color{blue}\textit{supported}} and {\color{red}\textit{unsupported}}, respectively.
Note that even when the description, "\textit{Dublin is Ireland’s national capital.}", \textit{is} factual (obtained from the perturbed fact), it is \textit{unsupported} by the original unperturbed fact, \textit{(Ireland, largest\_city, Dublin)}, and our HVM is trained to capture such faithfulness errors.
}
\label{tab:fate_hypos_examples}
\end{table}

To train the task-specific HVM with our FATE dataset, for each training instance we randomly set a decoding position and break its original and perturbed descriptions in two parts to simulate possible backward and forward hypotheses: a hypothesis derived from a \textit{perturbed} description that overlaps with the marked perturbed span receives \textit{unsupported} label as the ground truth, and all of the others receive \textit{supported}. We present an example synthetic pair of backward and forward hypotheses in Table~\ref{tab:fate_hypos_examples}. Finally, we up-sample the supported hypotheses to balance the labels.

% ========================================

\section{Experiments and Results}
\label{sec:expt}

% ------

% \subsection{Experiment Setup}

\noindent \textbf{Datasets and Models.} We train two base generation models BART-large~\cite{lewis-etal-2020-bart} and T5-large~\cite{2020t5},
following the hyperparameter settings from~\cite{ribeiro-etal-2021-investigating-LM-G2T-Generation}, and evaluate our decoding strategy on WebNLG~\cite{gardent-etal-2017-webnlg}, TekGen~\cite{agarwal-etal-2021-TekGen}, and GenWiki~\cite{jin-etal-2020-genwiki} 
% (see Appendix~\ref{appendix:eval-dataset-stats} for the dataset statistics)
.

\noindent \textbf{Metrics.} We assess the models on \textit{faithfulness} and \textit{quality}. Faithfulness metrics measure how much semantic distortion the output contains with respect to the input, while quality metrics measure how close a model output is to the reference. For the former we employ FactKB~\cite{feng2023factkb}, a state-of-the-art reference-free metric constructed via factuality pre-training. For the latter we employ the three metrics previously used by~\newcite{ribeiro-etal-2021-investigating-LM-G2T-Generation}: BLEU \cite{papineni-etal-2002-bleu}, METEOR \cite{banerjee-lavie-2005-meteor}, and BERTScore \cite{bert-score}.

\noindent \textbf{Baseline Decoding Strategies.} As baselines we test two basic decoding strategies: greedy search and beam search (Sec.~\ref{sec:task_defn}). For our TWEAK decoding strategy, we first test it with an off-the-shelf NLI model~\cite{nie-etal-2020-adversarial-nli} for hypothesis verification. Three variations are tested: TWEAK-NLI-B, TWEAK-NLI-F, and TWEAK-NLI-B+F, using only backward, only forward, and both hypotheses, respectively. We then replace the NLI model with our task-specific HVM trained with FATE dataset (Sec.~\ref{subsec:fate} \&~\ref{subsec:hvm}) as TWEAK-HVM variant. More implementation details are in Appendix~\ref{appendix:expt_details}. 

% ------

\subsection{Main Results in WebNLG}
\label{subsec:main_results}

%\begin{table*}[]
\begin{table}
\centering
\resizebox{\linewidth}{!}{
\begin{tabular}{cl|c|rrrr}
\toprule
%\multicolumn{1}{l}{\textbf{Model}} & \textbf{Decoding} & \textbf{FactKB} & \textbf{BLEU}  & \textbf{METEOR} & \textbf{BERTScore} \\ \midrule
\multicolumn{1}{l}{\textbf{}} & \textbf{Decoding} & \textbf{FKB} & \textbf{BLEU}  & \textbf{MET} & \textbf{BS} \\ \midrule
\multirow{6}{*}{\rotatebox{90}{BART-large}}              & Greedy            & 27.74           & 51.3           & 66.79           & 94.2               \\
                                   & Beam              & \textbf{28.91}           & \textbf{54.23} & \textbf{67.55}  & \textbf{94.35}     \\ \cmidrule{2-6} 
                                   & TWEAK-NLI-F       & 30.46           & {52.02}          & {67.17}           & {94.2}               \\
                                   & TWEAK-NLI-B       & {30.59}           & 49.68          & 65.88           & 94.12              \\
                                   & TWEAK-NLI-B+F     & 30.47           & 51.62          & 66.84           & 94.19              \\ \cmidrule{2-6} 
                                   & TWEAK-HVM         & \textbf{31.34}  & \textbf{53.14} & \textbf{67.38}  & \textbf{94.25}     \\ \midrule
\multirow{6}{*}{\rotatebox{90}{T5-large}}                & Greedy            & 30.14           & 57.71          & 68.71           & 94.84              \\
                                   & Beam              & \textbf{31.29}           & \textbf{58.93} & \textbf{69.38}           & \textbf{94.86}              \\ \cmidrule{2-6} 
                                   & TWEAK-NLI-F       & {33.03}           & {53.51}          & {67.8}            & {94.39}              \\
                                   & TWEAK-NLI-B       & 31.49           & 44.96          & 65.02           & 93.93              \\
                                   & TWEAK-NLI-B+F     & 32.71           & 51.71          & 66.73           & 94.19              \\ \cmidrule{2-6} 
                                   & TWEAK-HVM         & \textbf{33.34}  & \textbf{57.31} & \textbf{69.02}  & \textbf{94.68}     \\ \bottomrule
\end{tabular}
}
\caption{Results of decoding baselines and our TWEAK decoding variants measured by faithfulness metric (\textbf{FKB} = FactKB) and quality metrics (\textbf{BLEU}, \textbf{MET} = METEOR, \textbf{BS} = BERTScore) on WebNLG dataset. Numbers in \textbf{bold} are the highest scores among the baselines (greedy and beam) or among the TWEAK variants.}
\label{tab:main_results}
\end{table}

Our main results are shown in Table~\ref{tab:main_results}. Overall the best TWEAK variants improve on average +2.24 points on faithfulness (FactKB), with only -0.14 points degradation in quality (BERTScore).

\begin{table*}[h!]
\centering
\resizebox{0.9\linewidth}{!}{
\begin{tabular}{cl|c|ccc|c|ccc}
\toprule
\multirow{2}{*}{\textbf{Model}} & \multicolumn{1}{c|}{\multirow{2}{*}{\textbf{Decoding}}} & \multicolumn{4}{c|}{\textbf{TekGen}}                                    & \multicolumn{4}{c}{\textbf{GenWiki}}                                    \\ \cmidrule{3-10} 
& \multicolumn{1}{c|}{} & \textbf{FactKB} & \textbf{BLEU}  & \textbf{METEOR} & \textbf{BS} & \textbf{FactKB} & \textbf{BLEU}  & \textbf{METEOR} & \textbf{BS} \\ \midrule
\multirow{6}{*}{\rotatebox{90}{BART-large}} & Greedy & 9.44 & \textbf{22.42} & \textbf{44.21}  & 90.32 & 13.69 & 30.31 & 60.53 & 90.71 \\
& Beam & \textbf{11.57} & 21.34 & 43.86  & \textbf{90.52} & \textbf{14.24} & \textbf{37.48} & \textbf{63.16} & \textbf{91.67} \\ \cmidrule{2-10} 
& TWEAK-NLI-F & 14.77 & 15.92 & 38.48  & 88.25 & 18.97 & 24.61 & 55 & 90.13 \\
& TWEAK-NLI-B & 12.8 & \textbf{20.22} & \textbf{42.62}  & \textbf{90.51} & 16.48 & \textbf{31.08} & \textbf{58.53} & \textbf{91.37} \\
& TWEAK-NLI-B+F & \textbf{15.2} & 17.57 & 38.79  & 88.57 & \textbf{19.51} & 25.54 & 56.02 & 90.29 \\ \cmidrule{2-10} 
& TWEAK-HVM & 13.24 & 19.26 & 40.48  & 88.65 & 15.72 & 29.52 & 56.17 & 90.54 \\ \midrule
\multirow{6}{*}{\rotatebox{90}{T5-large}} & Greedy & 9.12 & 21.09 & \textbf{43.09}  & 90.52 & 14.22 & 30.45 & 58.89 & 90.54 \\
& Beam & \textbf{11.64} & \textbf{21.35} & 42.97  & \textbf{90.61} & \textbf{14.67} & \textbf{37.25} & \textbf{61.4} & \textbf{91.57} \\ \cmidrule{2-10}
& TWEAK-NLI-F & \textbf{16.51} & 8.57 & 37.48  & 87.88 & 25.22 & 12.65 & 50.47 & 88.8 \\
& TWEAK-NLI-B & 12.12 & 19.98 & 41.32  & \textbf{90.33} & 23.78 & 18.25 & 54.11 & 90.31 \\
& TWEAK-NLI-B+F & 15.86 & 10.66 & 38.44  & 88.55 & \textbf{29.57} & 11.53 & 49.58 & 88.49 \\ \cmidrule{2-10} 
& TWEAK-HVM & 13.44 & \textbf{21.51} & \textbf{41.61}  & 89.56 & 17.54 & \textbf{30.62} & \textbf{57.54} & \textbf{90.88} \\ \bottomrule
\end{tabular}
}
\caption{Generalization results on \textit{out-of-distribution} (OOD) test sets TekGen and GenWiki. \textbf{BS} = BERTScore. Numbers in \textbf{bold} are the highest scores among the baselines (greedy and beam) or among the TWEAK variants.}
\label{tab:ood_results}
\end{table*}

\noindent \textbf{Baseline Decoding Results.} Looking at the results of the two baseline decoding strategies, we observe that beam search consistently outperforms greedy search on both faithfulness and quality metrics. This suggests that increasing the beam size during decoding widens the exploration and generates a more faithful and higher quality output.

\noindent \textbf{TWEAK Decoding with NLI.} Comparing our TWEAK-NLI variants to the baselines, we find that all of them outperform beam search on faithfulness (FactKB), with TWEAK-NLI-B on BART-large improving +1.68 points, and TWEAK-NLI-F on T5-large improving +1.74 points over beam search. This demonstrates the effectiveness of performing hypothesis verification during decoding to improve output faithfulness. For each generator, a different variant achieves the best faithfulness result while the combo approach, TWEAK-NLI-B+F, is always in the middle. This indicates that simply combining the scores obtained from both hypotheses does not guarantee an optimal gain in faithfulness.

On the quality front, all TWEAK-NLI variants score lower on all metrics, with TWEAK-NLI-F showing the least regression. A manual analysis reveals that the more faithful generations exhibit a higher divergence from the reference (see Appendix~\ref{appendix:example} for an example). This is also consistent with ~\newcite{wan-etal-2023-faithfulness} who show that optimizing faithfulness can lead to lower textual similarity with reference. We also note that since quality metrics require reference while the faithful metric does not, any noise present in the reference may lead to a lower score even if the output is reasonable.

\noindent \textbf{TWEAK Decoding with HVM.} Comparing the TWEAK-HVM variant (Sec.~\ref{subsec:hvm}) to the baselines, TWEAK-HVM significantly outperforms in faithfulness: its FactKB score reaches 31.34 (+2.43 points improvement) and 33.34 (+2.05 points) on BART-large and T5-large over beam search, respectively. TWEAK-HVM is also more faithful than the most faithful TWEAK-NLI variant, demonstrating the advantage of a task-specific HVM and the benefits of performing triple-specific entailment classification. 

On output quality, TWEAK-HVM still fares lower than beam search, but it scores higher than all TWEAK-NLI variants on all metrics, therefore significantly closing the gaps to be almost on par with beam search, with only 0.1/0.18 decline in BERTScore for BART/T5, respectively. In summary, TWEAK-HVM is more faithful than the baselines with almost as good quality.

% ------

\begin{table}[t!]
\centering
\resizebox{0.99\linewidth}{!}{
%\begin{tabular}{c|p{0.09\textwidth}|p{0.09\textwidth}|p{0.12\textwidth}}
\begin{tabular}{c|c|c|c}
\toprule
              & \textbf{Faithfulness} & \textbf{Completeness} & \textbf{Readability} \\ \midrule
\textbf{NLI vs Beam} & 56.06\%           & \textbf{56.67\%}         & 36.07\%        \\
\textbf{HVM vs Beam}   & \textbf{59.09\%}           & 56.06\%         & \textbf{45.83\%}     \\
\bottomrule
\end{tabular}
}
\caption{Human evaluation on NLI vs. Beam and HVM vs. Beam. Numbers are win-rates over the non-similar output. Highest numbers in each aspect are bolded.}
\label{tab:human_eval}
\end{table}

\subsection{Out-of-distribution Evaluation}
\label{subsec:ood_results}

We have demonstrated that performing hypothesis verification during decoding can significantly enhance faithfulness without losing much of the overall quality on an \textit{in-distribution} (ID) test set.
% \footnote{Recall our task-specific HVM is trained on FATE which is based on WebNLG. The results in Sec.~\ref{subsec:main_results} are obtained on WebNLG test set.}
To evaluate the \textit{out-of-distribution} (OOD) effectiveness of our approach, we conducted experiments on two additional datasets that the HVM is not trained on: TekGen~\cite{agarwal-etal-2021-TekGen} and GenWiki~\cite{jin-etal-2020-genwiki}. 
We show the results for BART and T5-large in Table~\ref{tab:ood_results}. Overall the best TWEAK variants improve BART and T5 on average +7.17 points on faithfulness (FactKB), with only -0.32 points degradation on quality (BERTScore). 

TWEAK-HVM still outperforms the best baseline (beam search) on \textit{faithfulness}, yielding an average relative improvement of 14.95\%/14.98\% on TekGen/GenWiki, respectively. However, the best NLI variant outperforms TWEAK-HVM on faithfulness by an average relative margin of 18.82\%/46.35\% on TekGen/GenWiki, respectively. Since the NLI model is trained with OOD datasets, it appears to be more generalizable than our task-specific HVM in the OOD setup, as expected.

On the quality front, all TWEAK variants score lower than the best baseline, similar to the ID setting (Sec.~\ref{subsec:main_results}).
% albeit with slightly larger differences than in the ID results, especially on GenWiki, which appears to be the noisier of the two OOD datasets. 
% Interestingly, 
if we compare NLI vs HVM by picking first the most faithful TWEAK-NLI variant, \textit{it always performs worse on quality than the TWEAK-HVM variant}. For example, on TekGen with BART-large, comparing TWEAK-NLI-B+F, which has the highest FactKB score among all NLI variants, to TWEAK-HVM using BLEU, the HVM variant outperforms by 1.69 absolute points. It appears TWEAK-HVM is able to strike a better balance between faithfulness and quality.

\begin{table}[t!]
\centering
\resizebox{0.9\linewidth}{!}{
\begin{tabular}{c|c|ccc}
\toprule
              & \textbf{FactKB} & \textbf{BLEU} & \textbf{MET} & \textbf{BS} \\ \midrule
\textbf{BART} & 30.47           & 51.62         & 66.84        & 94.19      \\
w/o DA        & -0.12           & -2.63         & -1.34        & -0.12      \\ \midrule
\textbf{T5}   & 32.71           & 51.71         & 66.73        & 94.19      \\
w/o DA        & -0.79           & -12.21        & -2.21        & -0.49      \\ \bottomrule
\end{tabular}
}
\caption{Effect of dynamic aggregation (DA) with TWEAK-NLI-B+F on WebNLG and BART-large. \textbf{MET} and \textbf{BS} stand for METEOR and BERTScore.}
\label{tab:ablation_on_DA}
\end{table}

\subsection{Human Evaluation}
\label{subsec:human_eval}

We also conduct human evaluation on WebNLG for the output of the beam search, TWEAK-NLI, and TWEAK-HVM decoding. The graders are asked to compare side-by-side NLI vs. Beam and HVM vs. Beam on three aspects: \textbf{faithfulness} (whether an output contains only claims supported by the input), \textbf{completeness} (whether an output captures all of the input), and \textbf{readability} (whether an output is grammatical and easy to understand), and can choose between four grades: \textit{better than}, \textit{similar to}, \textit{worse than}, and \textit{can't decide}. We use T5-large as the base model, and sample the output uniformly across different numbers of input facts (1 to 7), resulting in 127 instances. The result is shown in Table~\ref{tab:human_eval} in terms of the win-rates over the output that are not marked as similar. Overall, consistent with the main results discussed in Sec.~\ref{subsec:main_results}, both TWEAK variants outperform the beam search baseline on faithfulness and completeness, but underperform on readability. In particular, TWEAK-HVM outperforms more than TWEAK-NLI on faithfulness and readability, with nearly identical completeness.

% ------

% ========================================

\section{Analysis}
\label{sec:analysis}

We report additional experiments and analyses in this section.

% ------

% \subsection{Dynamic Aggregation}
\label{subsec:da}

\noindent \textbf{Dynamic Aggregation.} As observed in Table~\ref{tab:main_results}, different models achieve peak faithfulness using \textit{either} backward \textit{or} forward hypotheses (BART favors backward while T5 favors forward). This implies both types of hypotheses can be useful in improving faithfulness of the output, which is borne out again by the OOD results reported in Table~\ref{tab:ood_results} where we observe that TWEAK-NLI-B+F, using both backward and forward hypotheses via \textit{dynamic aggregation} (DA; see Sec.~\ref{subsubsec:hv_nli}), becomes the most faithful variant. To assess DA's impact, we examine TWEAK-NLI-B+F without DA on WebNLG in Table~\ref{tab:ablation_on_DA}, revealing a clear performance drop in both faithfulness and quality. This underscores the importance of adapting weights placed on forward/backward hypotheses throughout the decoding process, as incomplete hypothesis verification can be less reliable.

% ------

\label{subsec:alpha}

\begin{figure}
    \centering
    \includegraphics[width=\linewidth]{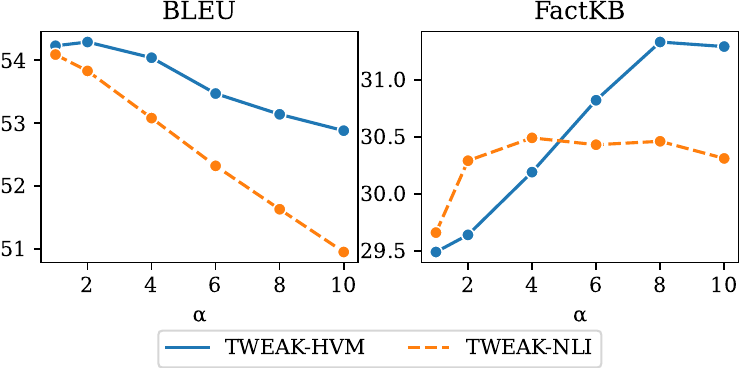}
    \caption{The effect on quality (BLEU) and faithfulness (FactKB) from choosing different $\alpha$ in Equ.~\eqref{equ:tweak_scoring}, with $\alpha = 0$ being equivalent to beam search. The results are obtained using TWEAK-NLI-B+F and TWEAK-HVM variants on WebNLG test set with BART.}
    \label{fig:weighting_effects}
\end{figure}

\noindent \textbf{Weighting Effects.} As described in Equ.~\eqref{equ:tweak_scoring}, we combine the generative score and the faithfulness score weighted by $\alpha$ to rank the candidates. We are therefore interested in the effect of choosing $\alpha$. In Fig.~\ref{fig:weighting_effects} we plot the resulting quality score (BLEU) and faithfulness score (FactKB) with different $\alpha$, with 0 being equivalent to beam search. The experiments are done with WebNLG test set and BART-large, using TWEAK-NLI-B+F and TWEAK-HVM variants.

We observe that increasing the weight on faithfulness score improves faithfulness in almost all settings at the cost of reduced quality. HVM outperforms NLI on quality at \textit{all} $\alpha$ values, and HVM also outperforms NLI on faithfulness when $\alpha \ge 6$. This clearly demonstrates the advantages of HVM in the ID setting (see Sec.~\ref{subsec:ood_results}). 

% ------

\label{subsec:num_input_facts}

\begin{table}[tb!]
\centering
%\resizebox{\linewidth}{!}{
{\small
\begin{tabular}{cllll}
\toprule
\multicolumn{2}{l}{\textbf{\#Triples}}   & \textbf{Short} & \textbf{Medium} & \textbf{Long} \\ \midrule
\multicolumn{2}{l}{\textbf{\#Sample}}      & 908            & 2196            & 620           \\ \midrule
\multirow{3}{*}{\rotatebox{90}{\textbf{BLEU}}}   & HVM         & 64.18          & 50.22           & 46.26         \\
                        & NLI         & 63.48          & 48.59           & 45.23         \\ \cmidrule{2-5}
                        & $\Delta$        & +1.09\%         & +3.25\%          & +2.23\%        \\ \midrule
\multirow{3}{*}{\rotatebox{90}{\textbf{FactKB}}} & HVM         & 18.11          & 33.47           & 43.17         \\
                        & NLI         & 18.06          & 32.67           & 40.81         \\ \cmidrule{2-5}
                        & $\Delta$        & +0.28\%         & +2.39\%          & +5.47\%        \\ \bottomrule
\end{tabular}
}
%}
\caption{TWEAK decoding performance on WebNLG with increasing number of input triples. We split the WebNLG test set into three groups: Short (1 triples), Medium (2-4 triples) and Long (5-7 triples).}
\label{tab:hvm_vs_nli_triples_length}
\end{table}

\noindent \textbf{Number of Input Facts.} The number of input fact triples is an important factor in determining K2T output quality: the more triples in the input, the more challenging for a model to generate a faithful and high-quality output. To investigate the correlation, we split the WebNLG test set into three groups: Short (one input triple), Medium (2-4 triples), and Long (5-7 triples). We then test both TWEAK-NLI-B+F and TWEAK-HVM variants with BART-large on these three groups. The results are shown in Table~\ref{tab:hvm_vs_nli_triples_length}.

On generative quality (BLEU) we observe that TWEAK-HVM outperforms TWEAK-NLI-B+F by a similar amount across the three groups. On faithfulness (FactKB), however, TWEAK-HVM's improvement over TWEAK-NLI-B+F is positively correlated with the number of input triples, climbing from +0.28\%, +2.39\%, to +5.47\% from Short, Medium, to Long. We attribute this growing advantage to HVM's ability to model each triple-hypothesis relation, whereas TWEAK-NLI-B+F concatenates all triples into a single premise and may misclassify with more triples in the input.

% ------

\label{subsec:beam_size}

\begin{figure}[tb!]
    \centering
    \includegraphics[width=0.9\linewidth]{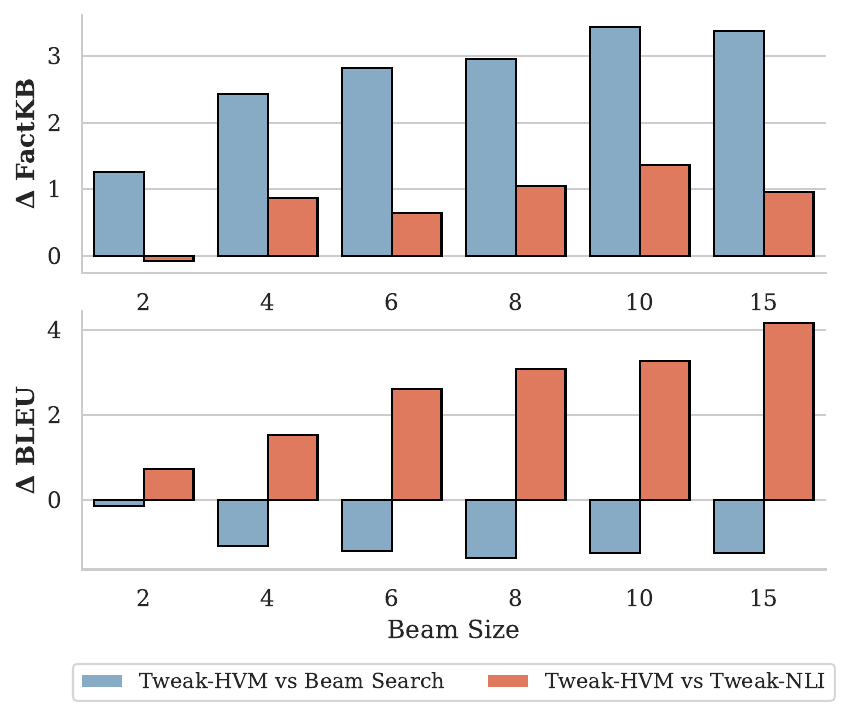}
    \caption{Performance differences ($\Delta$) on quality (BLEU) and faithfulness (FactKB) between TWEAK-HVM, TWEAK-NLI-B+F and beam search on various beam sizes $\{2,4,6,8,10,15\}$. All experiments are done on WebNLG with BART-large.}
    \label{fig:beam_size}
\end{figure}

\begin{figure}[tb!]
    \centering
    \includegraphics[width=0.9\linewidth]{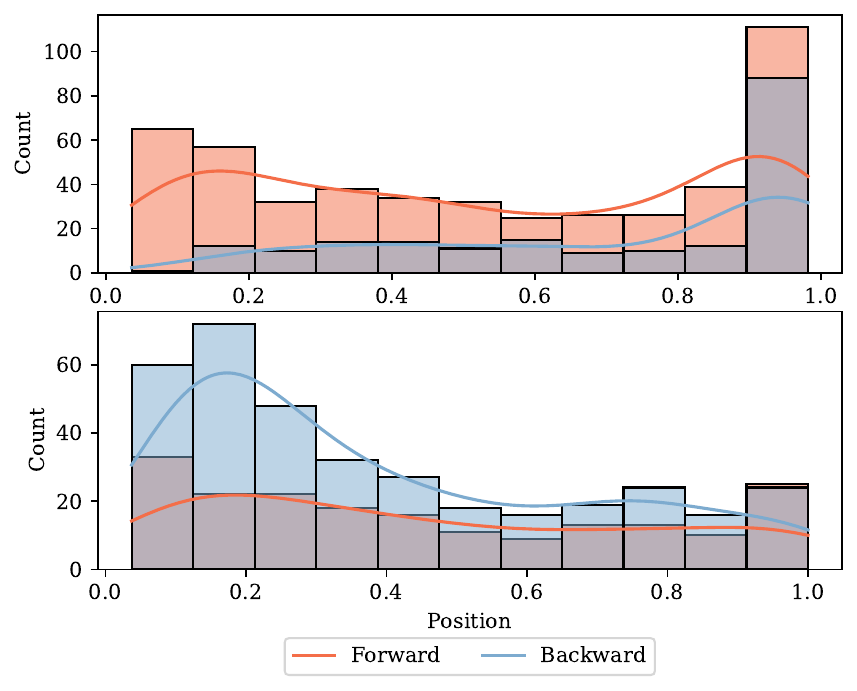}
    \caption{The distributions of the relative positions where negative predictions (i.e., possible hallucination) happen during the decoding process. $0$ and $1$ along the horizontal axis represent the start and end of the decoding. The upper and bottom panel represent TWEAK-HVM and TWEAK-NLI-B+F running on WebNLG with BART-large, respectively.}
    \label{fig:neg_relative_location}
\end{figure}

\noindent \textbf{Exploring Larger Beam Size.} If our TWEAK decoding strategy can promote a lower-ranked candidate based on its faithfulness score, can we further improve its effectiveness by increasing the beam size, i.e., letting in more candidates to be evaluated by TWEAK? To answer this question, we run beam search, TWEAK-NLI-B+F, and TWEAK-HVM side-by-side on WebNLG test set and BART-large, and plot their quality (BLEU) and faithfulness (FactKB) \textit{differences} in Fig.~\ref{fig:beam_size}.

Comparing TWEAK-HVM with beam search (blue bars), we observe that TWEAK-HVM improves on faithfulness, with improvement growing with beam size. 
% This growth is \textit{in addition} to beam search's own faithfulness improvement with increasing beam size reported in~\cite{wan-etal-2023-faithfulness}. 
In terms of quality, however, TWEAK-HVM underperforms beam search, but the drop stabilizes after beam size = 4.

Comparing TWEAK-HVM with TWEAK-NLI-B+F (red bars), we observe that on quality, TWEAK-HVM steadily becomes better than TWEAK-NLI-B+F as beam size increases. On faithfulness, TWEAK-HVM starts out being slightly worse at beam size = 2, but then steadily becomes better over TWEAK-NLI-B+F with increasing beam size until it reaches 10. This result shows TWEAK-HVM has a greater capacity in taking advantage of a bigger beam size.

% ------

\label{subsec:hallucination_pos}

\begin{figure*}[ht!]
    \centering
    \includegraphics[width=0.9\linewidth]{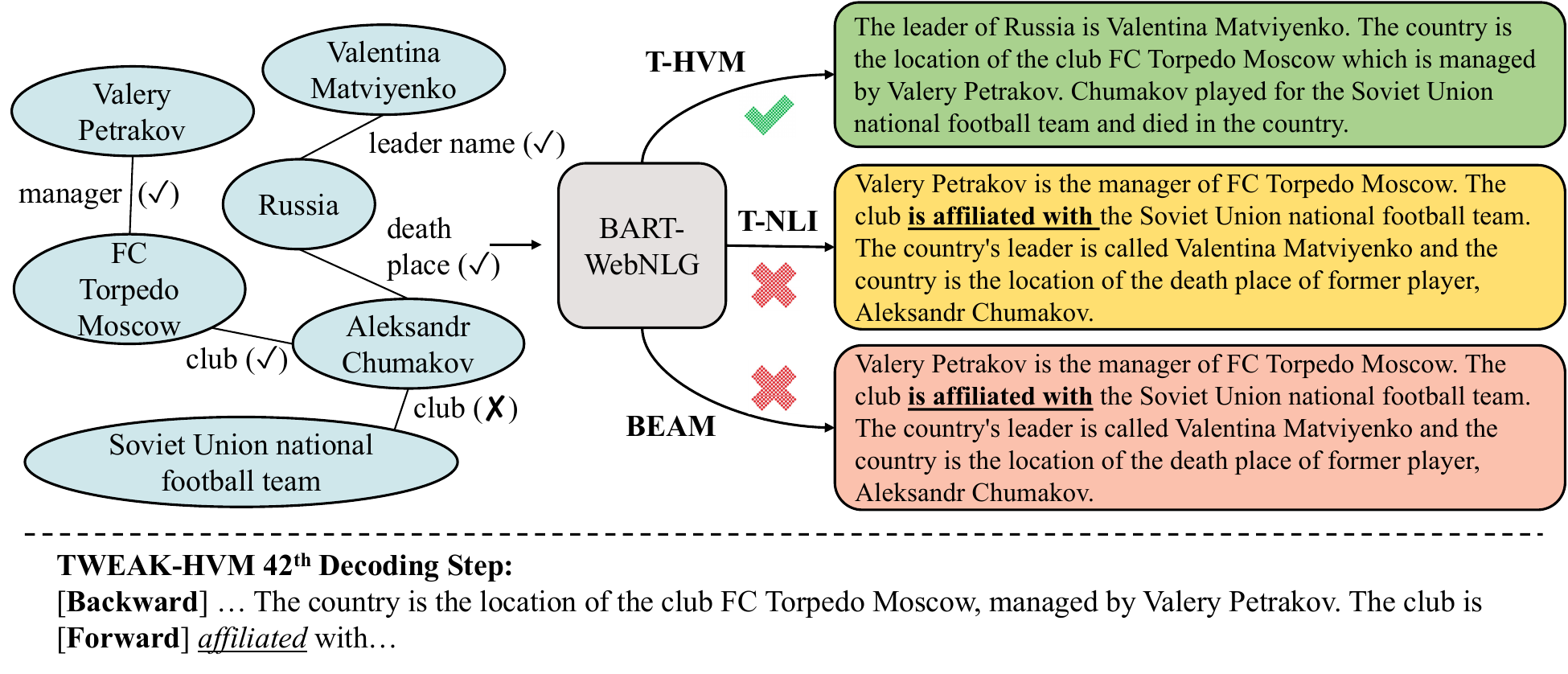}
    \caption{Output from beam search, TWEAK-NLI-B+F (T-NLI), and TWEAK-HVM (T-HVM) on an example taken from WebNLG test set, using BART-large. Tweak-HVM benefits from the more fine-grained modeling of hypothesis-triple relation and correctly capture the contradiction between forward hypothesis \textit{affiliated with...} and triple {\small \texttt{(Aleksandr Chumakov, club, Soviet Union national football team)}}. We use \cmark and \xmark to indicate HVM's predictions for triple-hypothesis pairs at the 40th decoding step.}
    \label{fig:qualitative_example}
\end{figure*}

\noindent \textbf{Where is Hallucination Found?} Since TWEAK's strength lies in its ability to identify and demote potential hallucinations at any decoding step, we are interested in investigating where these hallucinations can typically be detected. We experiment with TWEAK-NLI-B+F and TWEAK-HVM on WebNLG and BART-large, and analyze the distribution of predicted hallucination positions, normalized between 0 (beginning) and 1 (end), for backward and forward hypotheses. As depicted in Fig.\ref{fig:neg_relative_location}, TWEAK-HVM predicts more hallucinating forward hypotheses, while TWEAK-NLI-B+F leans towards more hallucinating backward hypotheses.

This divergence can be attributed to the training differences between NLI and HVM. NLI, trained on complete hypotheses, tends to assign lower entailment scores to incomplete sentences like backward hypotheses, leading to negative predictions in the $\nli$ function used in Equ.\eqref{equ:hv_nli}. In contrast, HVM's inclination towards forward hallucinations might stem from FATE's uneven perturbation distribution, where objects and relations undergo the most perturbations. Since objects and relations are predominantly positioned toward sentence endings in a Subject-Verb-Object language like English, the trained HVM may detect more forward hallucinations due to their higher likelihood of containing perturbations in the training set. When amplified by dynamic aggregation which gives early emphasis to forward hypotheses, this helps stop potential errors from happening earlier in decoding. This also explains why TWEAK-HVM rarely detects backward hallucinations at the start, and why TWEAK-NLI-B+F initially detects more hallucinations.

% This rationale also clarifies why TWEAK-HVM rarely detects backward hallucinations at the start (as shorter sentence prefixes are unlikely to contain perturbations typical in later segments of FATE) and why TWEAK-NLI-B+F initially detects more hallucinations (attributed to the extreme incompleteness of backward hypotheses early on). These insights highlight a potential avenue for future research: achieving a more balanced triple perturbation distribution when constructing FATE.

\noindent \textbf{Qualitative Case.} We offer an example in Fig.~\ref{fig:qualitative_example} that shows how TWEAK-HVM successfully directs the decoding process away from a potential hallucination. The example features five input fact triples describing the professional relationships around footballer Aleksandr Chumakov. Both beam search and TWEAK-NLI produced hallucinating output, describing that ``\textit{FC Torpedo Moscow}'' is affiliated with ``\textit{the Soviet Union national football team}'', which is \textit{not} stated in the input facts. The hallucination stems from the wrong interpretation of triple \texttt{\small{(Aleksandr Chumakov, club, Soviet Union national football team)}}, which TWEAK-HVM correctly concludes is not supported by the forward hypothesis ``\textit{affiliated with...}'' at the 40th decoding step. More examples can be found in Appendix~\ref{appendix:more-qualitative-examples}.

% ------

% ========================================

\section{Conclusions}
\label{sec:conclusions}

We introduce TWEAK, a model-agnostic decoding strategy incorporating hypothesis verification, to mitigate hallucinations in K2T generation. 
% Addressing the limitations of autoregressive generation, 
% TWEAK evaluates candidates based on their alignment with input facts, significantly enhancing output faithfulness. 
Our work demonstrates the effectiveness of TWEAK with an off-the-shelf NLI model and a task-specific HVM. Future directions involve improving generalization,
% refining our knowledge perturbation method and 
and reducing inference costs via techniques such as knowledge distillation \cite{wan-etal-2023-faithfulness}.

\section*{Ethical Considerations}
\label{sec:ethical-consideration}

This paper focuses on accurate knowledge-to-text generation, crucial for reducing errors in natural language generation. Our goal is to minimize mistakes and misinformation synthesized in texts produced by the generative language models. Within this scope, TWEAK provides a method to manipulate text generation to be more faithful without altering the trained generative model.

In our human evaluation, all participants are employed as full-time workers within our internal data annotation team. To guarantee evaluation quality, we ensure that all participants possess a native-speaker level of proficiency in English. Each participant receives fair compensation, commensurate with standard wages in the United States.
All participants are explicitly informed that the annotated data would be used for research purposes.
Additionally, this study has undergone review and approval by our internal ethical panel.

\section*{Limitations}
\label{sec:limitations}

The authors wish to note the following limitations:

\begin{itemizesquish}{-0.3em}{0.5em}
    \item The proposed TWEAK decoding strategy imposes additional cost at inference time compared to the baseline approaches such as beam search.
    \item The reported results indicate while all TWEAK variants outperform the baseline in ID settings, in OOD settings the results are more nuanced. On faithfulness, TWEAK-HVM still outperforms the baselines in OOD settings, but it underperforms the more costly variant TWEAK-NLI in some settings (see Sec.~\ref{subsec:ood_results}). A future exploration is to further improve the robustness of HVM as discussed in Sec.~\ref{subsec:hallucination_pos}.
    \item Our proposed approach has only been tested in English language. The authors expect the approach to work reasonably well in non-English languages, provided adequate datasets and base models are available.
\end{itemizesquish}

\section*{Acknowledgements}
We thank the reviewers for their useful feedback. We are grateful for the 2023 Apple Scholarship awarded to Yifu Qiu. 

% Entries for the entire Anthology, followed by custom entries
\bibliography{anthology,custom}

\begin{thebibliography}{52}
\expandafter\ifx\csname natexlab\endcsname\relax\def\natexlab#1{#1}\fi

\bibitem[{Agarwal et~al.(2021)Agarwal, Ge, Shakeri, and Al-Rfou}]{agarwal-etal-2021-TekGen}
Oshin Agarwal, Heming Ge, Siamak Shakeri, and Rami Al-Rfou. 2021.
\newblock \href {https://doi.org/10.18653/v1/2021.naacl-main.278} {Knowledge graph based synthetic corpus generation for knowledge-enhanced language model pre-training}.
\newblock In \emph{Proceedings of the 2021 Conference of the North American Chapter of the Association for Computational Linguistics: Human Language Technologies}, pages 3554--3565, Online. Association for Computational Linguistics.

\bibitem[{Banerjee and Lavie(2005)}]{banerjee-lavie-2005-meteor}
Satanjeev Banerjee and Alon Lavie. 2005.
\newblock \href {https://aclanthology.org/W05-0909} {{METEOR}: An automatic metric for {MT} evaluation with improved correlation with human judgments}.
\newblock In \emph{Proceedings of the {ACL} Workshop on Intrinsic and Extrinsic Evaluation Measures for Machine Translation and/or Summarization}, pages 65--72, Ann Arbor, Michigan. Association for Computational Linguistics.

\bibitem[{Chen et~al.(2022)Chen, Lu, Xu, Li, Jingbo, Dou, and Xiong}]{chen-etal-2022-towards-table2text-generation}
Miao Chen, Xinjiang Lu, Tong Xu, Yanyan Li, Zhou Jingbo, Dejing Dou, and Hui Xiong. 2022.
\newblock \href {https://doi.org/10.18653/v1/2022.emnlp-main.562} {Towards table-to-text generation with pretrained language model: A table structure understanding and text deliberating approach}.
\newblock In \emph{Proceedings of the 2022 Conference on Empirical Methods in Natural Language Processing}, pages 8199--8210, Abu Dhabi, United Arab Emirates. Association for Computational Linguistics.

\bibitem[{Colas et~al.(2021)Colas, Sadeghian, Wang, and Wang}]{colas2021eventnarrative}
Anthony Colas, Ali Sadeghian, Yue Wang, and Daisy~Zhe Wang. 2021.
\newblock Eventnarrative: A large-scale event-centric dataset for knowledge graph-to-text generation.
\newblock In \emph{Thirty-fifth Conference on Neural Information Processing Systems Datasets and Benchmarks Track (Round 1)}.

\bibitem[{Daheim et~al.(2023)Daheim, Dziri, Sachan, Gurevych, and Ponti}]{daheim2023elastic}
Nico Daheim, Nouha Dziri, Mrinmaya Sachan, Iryna Gurevych, and Edoardo~M. Ponti. 2023.
\newblock \href {http://arxiv.org/abs/2303.17574} {Elastic weight removal for faithful and abstractive dialogue generation}.

\bibitem[{Dziri et~al.(2022)Dziri, Kamalloo, Milton, Zaiane, Yu, Ponti, and Reddy}]{dziri-etal-2022-faithdial}
Nouha Dziri, Ehsan Kamalloo, Sivan Milton, Osmar Zaiane, Mo~Yu, Edoardo~M. Ponti, and Siva Reddy. 2022.
\newblock \href {https://doi.org/10.1162/tacl_a_00529} {{F}aith{D}ial: A faithful benchmark for information-seeking dialogue}.
\newblock \emph{Transactions of the Association for Computational Linguistics}, 10:1473--1490.

\bibitem[{Fatahi~Bayat et~al.(2022)Fatahi~Bayat, Bhutani, and Jagadish}]{fatahi-bayat-etal-2022-compactie}
Farima Fatahi~Bayat, Nikita Bhutani, and H.~Jagadish. 2022.
\newblock \href {https://doi.org/10.18653/v1/2022.naacl-main.65} {{C}ompact{IE}: Compact facts in open information extraction}.
\newblock In \emph{Proceedings of the 2022 Conference of the North American Chapter of the Association for Computational Linguistics: Human Language Technologies}, pages 900--910, Seattle, United States. Association for Computational Linguistics.

\bibitem[{Feng et~al.(2023)Feng, Balachandran, Bai, and Tsvetkov}]{feng2023factkb}
Shangbin Feng, Vidhisha Balachandran, Yuyang Bai, and Yulia Tsvetkov. 2023.
\newblock Factkb: Generalizable factuality evaluation using language models enhanced with factual knowledge.
\newblock \emph{arXiv preprint arXiv:2305.08281}.

\bibitem[{Gardent et~al.(2017)Gardent, Shimorina, Narayan, and Perez-Beltrachini}]{gardent-etal-2017-webnlg}
Claire Gardent, Anastasia Shimorina, Shashi Narayan, and Laura Perez-Beltrachini. 2017.
\newblock \href {https://doi.org/10.18653/v1/W17-3518} {The {W}eb{NLG} challenge: Generating text from {RDF} data}.
\newblock In \emph{Proceedings of the 10th International Conference on Natural Language Generation}, pages 124--133, Santiago de Compostela, Spain. Association for Computational Linguistics.

\bibitem[{Guo et~al.(2019)Guo, Zhang, Teng, and Lu}]{guo-etal-2019-densely}
Zhijiang Guo, Yan Zhang, Zhiyang Teng, and Wei Lu. 2019.
\newblock \href {https://doi.org/10.1162/tacl_a_00269} {Densely connected graph convolutional networks for graph-to-sequence learning}.
\newblock \emph{Transactions of the Association for Computational Linguistics}, 7:297--312.

\bibitem[{Han and Shareghi(2022)}]{han-shareghi-2022-self}
Jiuzhou Han and Ehsan Shareghi. 2022.
\newblock \href {https://aclanthology.org/2022.emnlp-main.321} {Self-supervised graph masking pre-training for graph-to-text generation}.
\newblock In \emph{Proceedings of the 2022 Conference on Empirical Methods in Natural Language Processing}, pages 4845--4853, Abu Dhabi, United Arab Emirates. Association for Computational Linguistics.

\bibitem[{Hashem et~al.(2023)Hashem, Wang, Wijaya, Ali, and Li}]{hashem-etal-2023-generating}
Tahsina Hashem, Weiqing Wang, Derry~Tanti Wijaya, Mohammed~Eunus Ali, and Yuan-Fang Li. 2023.
\newblock \href {https://aclanthology.org/2023.inlg-main.8} {Generating faithful text from a knowledge graph with noisy reference text}.
\newblock In \emph{Proceedings of the 16th International Natural Language Generation Conference}, pages 106--122, Prague, Czechia. Association for Computational Linguistics.

\bibitem[{Ji et~al.(2023)Ji, Liu, Lee, Yu, Wilie, Zeng, and Fung}]{ji-etal-2023-rho}
Ziwei Ji, Zihan Liu, Nayeon Lee, Tiezheng Yu, Bryan Wilie, Min Zeng, and Pascale Fung. 2023.
\newblock \href {https://aclanthology.org/2023.findings-acl.275} {{RHO}: Reducing hallucination in open-domain dialogues with knowledge grounding}.
\newblock In \emph{Findings of the Association for Computational Linguistics: ACL 2023}, pages 4504--4522, Toronto, Canada. Association for Computational Linguistics.

\bibitem[{Jin et~al.(2020)Jin, Guo, Qiu, and Zhang}]{jin-etal-2020-genwiki}
Zhijing Jin, Qipeng Guo, Xipeng Qiu, and Zheng Zhang. 2020.
\newblock \href {https://doi.org/10.18653/v1/2020.coling-main.217} {{G}en{W}iki: A dataset of 1.3 million content-sharing text and graphs for unsupervised graph-to-text generation}.
\newblock In \emph{Proceedings of the 28th International Conference on Computational Linguistics}, pages 2398--2409, Barcelona, Spain (Online). International Committee on Computational Linguistics.

\bibitem[{Koncel-Kedziorski et~al.(2019)Koncel-Kedziorski, Bekal, Luan, Lapata, and Hajishirzi}]{koncel-kedziorski-etal-2019-text}
Rik Koncel-Kedziorski, Dhanush Bekal, Yi~Luan, Mirella Lapata, and Hannaneh Hajishirzi. 2019.
\newblock \href {https://doi.org/10.18653/v1/N19-1238} {{T}ext {G}eneration from {K}nowledge {G}raphs with {G}raph {T}ransformers}.
\newblock In \emph{Proceedings of the 2019 Conference of the North {A}merican Chapter of the Association for Computational Linguistics: Human Language Technologies, Volume 1 (Long and Short Papers)}, pages 2284--2293, Minneapolis, Minnesota. Association for Computational Linguistics.

\bibitem[{Kryscinski et~al.(2020)Kryscinski, McCann, Xiong, and Socher}]{kryscinski-etal-2020-FactCC}
Wojciech Kryscinski, Bryan McCann, Caiming Xiong, and Richard Socher. 2020.
\newblock \href {https://doi.org/10.18653/v1/2020.emnlp-main.750} {Evaluating the factual consistency of abstractive text summarization}.
\newblock In \emph{Proceedings of the 2020 Conference on Empirical Methods in Natural Language Processing (EMNLP)}, pages 9332--9346, Online. Association for Computational Linguistics.

\bibitem[{Laban et~al.(2022)Laban, Schnabel, Bennett, and Hearst}]{laban-etal-2022-summac}
Philippe Laban, Tobias Schnabel, Paul~N. Bennett, and Marti~A. Hearst. 2022.
\newblock \href {https://doi.org/10.1162/tacl_a_00453} {{S}umma{C}: Re-visiting {NLI}-based models for inconsistency detection in summarization}.
\newblock \emph{Transactions of the Association for Computational Linguistics}, 10:163--177.

\bibitem[{Lewis et~al.(2020)Lewis, Liu, Goyal, Ghazvininejad, Mohamed, Levy, Stoyanov, and Zettlemoyer}]{lewis-etal-2020-bart}
Mike Lewis, Yinhan Liu, Naman Goyal, Marjan Ghazvininejad, Abdelrahman Mohamed, Omer Levy, Veselin Stoyanov, and Luke Zettlemoyer. 2020.
\newblock \href {https://doi.org/10.18653/v1/2020.acl-main.703} {{BART}: Denoising sequence-to-sequence pre-training for natural language generation, translation, and comprehension}.
\newblock In \emph{Proceedings of the 58th Annual Meeting of the Association for Computational Linguistics}, pages 7871--7880, Online. Association for Computational Linguistics.

\bibitem[{Li et~al.(2021)Li, Tang, Zhao, Wei, Yuan, and Wen}]{li-etal-2021-shot-knowledge}
Junyi Li, Tianyi Tang, Wayne~Xin Zhao, Zhicheng Wei, Nicholas~Jing Yuan, and Ji-Rong Wen. 2021.
\newblock \href {https://doi.org/10.18653/v1/2021.findings-acl.136} {Few-shot knowledge graph-to-text generation with pretrained language models}.
\newblock In \emph{Findings of the Association for Computational Linguistics: ACL-IJCNLP 2021}, pages 1558--1568, Online. Association for Computational Linguistics.

\bibitem[{Liu et~al.(2019)Liu, Ott, Goyal, Du, Joshi, Chen, Levy, Lewis, Zettlemoyer, and Stoyanov}]{liu2019roberta}
Yinhan Liu, Myle Ott, Naman Goyal, Jingfei Du, Mandar Joshi, Danqi Chen, Omer Levy, Mike Lewis, Luke Zettlemoyer, and Veselin Stoyanov. 2019.
\newblock \href {http://arxiv.org/abs/1907.11692} {Roberta: A robustly optimized bert pretraining approach}.

\bibitem[{Liu et~al.(2022)Liu, Liu, Radev, and Neubig}]{liu-etal-2022-brio-summ}
Yixin Liu, Pengfei Liu, Dragomir Radev, and Graham Neubig. 2022.
\newblock \href {https://doi.org/10.18653/v1/2022.acl-long.207} {{BRIO}: Bringing order to abstractive summarization}.
\newblock In \emph{Proceedings of the 60th Annual Meeting of the Association for Computational Linguistics (Volume 1: Long Papers)}, pages 2890--2903, Dublin, Ireland. Association for Computational Linguistics.

\bibitem[{Lu et~al.(2022)Lu, Welleck, West, Jiang, Kasai, Khashabi, Le~Bras, Qin, Yu, Zellers, Smith, and Choi}]{lu-etal-2022-neurologic-astar}
Ximing Lu, Sean Welleck, Peter West, Liwei Jiang, Jungo Kasai, Daniel Khashabi, Ronan Le~Bras, Lianhui Qin, Youngjae Yu, Rowan Zellers, Noah~A. Smith, and Yejin Choi. 2022.
\newblock \href {https://doi.org/10.18653/v1/2022.naacl-main.57} {{N}euro{L}ogic a*esque decoding: Constrained text generation with lookahead heuristics}.
\newblock In \emph{Proceedings of the 2022 Conference of the North American Chapter of the Association for Computational Linguistics: Human Language Technologies}, pages 780--799, Seattle, United States. Association for Computational Linguistics.

\bibitem[{Marcheggiani and Perez-Beltrachini(2018)}]{marcheggiani-perez-beltrachini-2018-deep}
Diego Marcheggiani and Laura Perez-Beltrachini. 2018.
\newblock \href {https://doi.org/10.18653/v1/W18-6501} {Deep graph convolutional encoders for structured data to text generation}.
\newblock In \emph{Proceedings of the 11th International Conference on Natural Language Generation}, pages 1--9, Tilburg University, The Netherlands. Association for Computational Linguistics.

\bibitem[{Maynez et~al.(2020)Maynez, Narayan, Bohnet, and McDonald}]{maynez-etal-2020-factuality-abs-summ}
Joshua Maynez, Shashi Narayan, Bernd Bohnet, and Ryan McDonald. 2020.
\newblock \href {https://doi.org/10.18653/v1/2020.acl-main.173} {On faithfulness and factuality in abstractive summarization}.
\newblock In \emph{Proceedings of the 58th Annual Meeting of the Association for Computational Linguistics}, pages 1906--1919, Online. Association for Computational Linguistics.

\bibitem[{Nie et~al.(2020)Nie, Williams, Dinan, Bansal, Weston, and Kiela}]{nie-etal-2020-adversarial-nli}
Yixin Nie, Adina Williams, Emily Dinan, Mohit Bansal, Jason Weston, and Douwe Kiela. 2020.
\newblock Adversarial {NLI}: A new benchmark for natural language understanding.
\newblock In \emph{Proceedings of the 58th Annual Meeting of the Association for Computational Linguistics}. Association for Computational Linguistics.

\bibitem[{Papineni et~al.(2002)Papineni, Roukos, Ward, and Zhu}]{papineni-etal-2002-bleu}
Kishore Papineni, Salim Roukos, Todd Ward, and Wei-Jing Zhu. 2002.
\newblock \href {https://doi.org/10.3115/1073083.1073135} {{B}leu: a method for automatic evaluation of machine translation}.
\newblock In \emph{Proceedings of the 40th Annual Meeting of the Association for Computational Linguistics}, pages 311--318, Philadelphia, Pennsylvania, USA. Association for Computational Linguistics.

\bibitem[{Perez-Beltrachini and Lapata(2018)}]{perez-lapata2018}
Laura Perez-Beltrachini and Mirella Lapata. 2018.
\newblock Bootstrapping generators from noisy data.
\newblock In \emph{North American Chapter of the Association for Computational Linguistics}, New Orleans, Louisiana. Association for Computational Linguistics.
\newblock (NAACL 2018).

\bibitem[{Puduppully et~al.(2019)Puduppully, Dong, and Lapata}]{10.1609/aaai.v33i01.33016908}
Ratish Puduppully, Li~Dong, and Mirella Lapata. 2019.
\newblock \href {https://doi.org/10.1609/aaai.v33i01.33016908} {Data-to-text generation with content selection and planning}.
\newblock In \emph{Proceedings of the Thirty-Third AAAI Conference on Artificial Intelligence and Thirty-First Innovative Applications of Artificial Intelligence Conference and Ninth AAAI Symposium on Educational Advances in Artificial Intelligence}, AAAI'19/IAAI'19/EAAI'19. AAAI Press.

\bibitem[{Puduppully et~al.(2022)Puduppully, Fu, and Lapata}]{10.1162/tacl_a_00484}
Ratish Puduppully, Yao Fu, and Mirella Lapata. 2022.
\newblock \href {https://doi.org/10.1162/tacl_a_00484} {{Data-to-text Generation with Variational Sequential Planning}}.
\newblock \emph{Transactions of the Association for Computational Linguistics}, 10:697--715.

\bibitem[{Puduppully and Lapata(2021)}]{10.1162/tacl_a_00381-marco-plan-generation}
Ratish Puduppully and Mirella Lapata. 2021.
\newblock \href {https://doi.org/10.1162/tacl_a_00381} {{Data-to-text Generation with Macro Planning}}.
\newblock \emph{Transactions of the Association for Computational Linguistics}, 9:510--527.

\bibitem[{Qiu and Cohen(2022)}]{qiu-cohen-2022-hiergnn}
Yifu Qiu and Shay~B. Cohen. 2022.
\newblock \href {https://doi.org/10.18653/v1/2022.emnlp-main.355} {Abstractive summarization guided by latent hierarchical document structure}.
\newblock In \emph{Proceedings of the 2022 Conference on Empirical Methods in Natural Language Processing}, pages 5303--5317, Abu Dhabi, United Arab Emirates. Association for Computational Linguistics.

\bibitem[{Qiu et~al.(2023)Qiu, Ziser, Korhonen, Ponti, and Cohen}]{qiu2023detecting-mfact}
Yifu Qiu, Yftah Ziser, Anna Korhonen, Edoardo~M. Ponti, and Shay~B. Cohen. 2023.
\newblock \href {http://arxiv.org/abs/2305.13632} {Detecting and mitigating hallucinations in multilingual summarisation}.

\bibitem[{Raffel et~al.(2020)Raffel, Shazeer, Roberts, Lee, Narang, Matena, Zhou, Li, and Liu}]{2020t5}
Colin Raffel, Noam Shazeer, Adam Roberts, Katherine Lee, Sharan Narang, Michael Matena, Yanqi Zhou, Wei Li, and Peter~J. Liu. 2020.
\newblock \href {http://jmlr.org/papers/v21/20-074.html} {Exploring the limits of transfer learning with a unified text-to-text transformer}.
\newblock \emph{Journal of Machine Learning Research}, 21(140):1--67.

\bibitem[{Rebuffel et~al.(2020)Rebuffel, Soulier, Scoutheeten, and Gallinari}]{rebuffel2020hierarchical}
Cl{\'e}ment Rebuffel, Laure Soulier, Geoffrey Scoutheeten, and Patrick Gallinari. 2020.
\newblock A hierarchical model for data-to-text generation.
\newblock In \emph{Advances in Information Retrieval: 42nd European Conference on IR Research, ECIR 2020, Lisbon, Portugal, April 14--17, 2020, Proceedings, Part I 42}, pages 65--80. Springer.

\bibitem[{Ribeiro et~al.(2021{\natexlab{a}})Ribeiro, Schmitt, Sch{\"u}tze, and Gurevych}]{ribeiro-etal-2021-investigating-LM-G2T-Generation}
Leonardo F.~R. Ribeiro, Martin Schmitt, Hinrich Sch{\"u}tze, and Iryna Gurevych. 2021{\natexlab{a}}.
\newblock \href {https://doi.org/10.18653/v1/2021.nlp4convai-1.20} {Investigating pretrained language models for graph-to-text generation}.
\newblock In \emph{Proceedings of the 3rd Workshop on Natural Language Processing for Conversational AI}, pages 211--227, Online. Association for Computational Linguistics.

\bibitem[{Ribeiro et~al.(2021{\natexlab{b}})Ribeiro, Schmitt, Sch{\"u}tze, and Gurevych}]{ribeiro-etal-2021-investigating-plm-for-graph-to-text-generation}
Leonardo F.~R. Ribeiro, Martin Schmitt, Hinrich Sch{\"u}tze, and Iryna Gurevych. 2021{\natexlab{b}}.
\newblock \href {https://doi.org/10.18653/v1/2021.nlp4convai-1.20} {Investigating pretrained language models for graph-to-text generation}.
\newblock In \emph{Proceedings of the 3rd Workshop on Natural Language Processing for Conversational AI}, pages 211--227, Online. Association for Computational Linguistics.

\bibitem[{Schmidt(2019)}]{schmidt-2019-generalization-exposurebias}
Florian Schmidt. 2019.
\newblock \href {https://doi.org/10.18653/v1/D19-5616} {Generalization in generation: A closer look at exposure bias}.
\newblock In \emph{Proceedings of the 3rd Workshop on Neural Generation and Translation}, pages 157--167, Hong Kong. Association for Computational Linguistics.

\bibitem[{Schmitt et~al.(2021)Schmitt, Ribeiro, Dufter, Gurevych, and Sch{\"u}tze}]{schmitt-etal-2021-modeling}
Martin Schmitt, Leonardo F.~R. Ribeiro, Philipp Dufter, Iryna Gurevych, and Hinrich Sch{\"u}tze. 2021.
\newblock \href {https://doi.org/10.18653/v1/2021.textgraphs-1.2} {Modeling graph structure via relative position for text generation from knowledge graphs}.
\newblock In \emph{Proceedings of the Fifteenth Workshop on Graph-Based Methods for Natural Language Processing (TextGraphs-15)}, pages 10--21, Mexico City, Mexico. Association for Computational Linguistics.

\bibitem[{Su et~al.(2021)Su, Meng, Baker, and Collier}]{su-etal-2021-shot-table}
Yixuan Su, Zaiqiao Meng, Simon Baker, and Nigel Collier. 2021.
\newblock \href {https://doi.org/10.18653/v1/2021.findings-emnlp.77} {Few-shot table-to-text generation with prototype memory}.
\newblock In \emph{Findings of the Association for Computational Linguistics: EMNLP 2021}, pages 910--917, Punta Cana, Dominican Republic. Association for Computational Linguistics.

\bibitem[{Utama et~al.(2022)Utama, Bambrick, Moosavi, and Gurevych}]{utama-etal-2022-falsesum}
Prasetya Utama, Joshua Bambrick, Nafise Moosavi, and Iryna Gurevych. 2022.
\newblock \href {https://doi.org/10.18653/v1/2022.naacl-main.199} {Falsesum: Generating document-level {NLI} examples for recognizing factual inconsistency in summarization}.
\newblock In \emph{Proceedings of the 2022 Conference of the North American Chapter of the Association for Computational Linguistics: Human Language Technologies}, pages 2763--2776, Seattle, United States. Association for Computational Linguistics.

\bibitem[{Wan et~al.(2023)Wan, Liu, McKeown, Dreyer, and Bansal}]{wan-etal-2023-faithfulness}
David Wan, Mengwen Liu, Kathleen McKeown, Markus Dreyer, and Mohit Bansal. 2023.
\newblock \href {https://aclanthology.org/2023.eacl-main.210} {Faithfulness-aware decoding strategies for abstractive summarization}.
\newblock In \emph{Proceedings of the 17th Conference of the European Chapter of the Association for Computational Linguistics}, pages 2864--2880, Dubrovnik, Croatia. Association for Computational Linguistics.

\bibitem[{Wang et~al.(2022)Wang, Xu, Szekely, and Chen}]{wang-etal-2022-robust}
Fei Wang, Zhewei Xu, Pedro Szekely, and Muhao Chen. 2022.
\newblock \href {https://doi.org/10.18653/v1/2022.naacl-main.371} {Robust (controlled) table-to-text generation with structure-aware equivariance learning}.
\newblock In \emph{Proceedings of the 2022 Conference of the North American Chapter of the Association for Computational Linguistics: Human Language Technologies}, pages 5037--5048, Seattle, United States. Association for Computational Linguistics.

\bibitem[{Wang et~al.(2021)Wang, Sun, Wu, Zhou, Li, and Yan}]{wang-etal-2021-unire-tableloss}
Yijun Wang, Changzhi Sun, Yuanbin Wu, Hao Zhou, Lei Li, and Junchi Yan. 2021.
\newblock \href {https://doi.org/10.18653/v1/2021.acl-long.19} {{U}ni{RE}: A unified label space for entity relation extraction}.
\newblock In \emph{Proceedings of the 59th Annual Meeting of the Association for Computational Linguistics and the 11th International Joint Conference on Natural Language Processing (Volume 1: Long Papers)}, pages 220--231, Online. Association for Computational Linguistics.

\bibitem[{Wang et~al.(2023)Wang, Collins, Vedula, Filice, Malmasi, and Rokhlenko}]{wang-etal-2023-faithful}
Zhuoer Wang, Marcus Collins, Nikhita Vedula, Simone Filice, Shervin Malmasi, and Oleg Rokhlenko. 2023.
\newblock \href {https://aclanthology.org/2023.acl-long.160} {Faithful low-resource data-to-text generation through cycle training}.
\newblock In \emph{Proceedings of the 61st Annual Meeting of the Association for Computational Linguistics (Volume 1: Long Papers)}, pages 2847--2867, Toronto, Canada. Association for Computational Linguistics.

\bibitem[{Wolf et~al.(2020)Wolf, Debut, Sanh, Chaumond, Delangue, Moi, Cistac, Rault, Louf, Funtowicz, Davison, Shleifer, von Platen, Ma, Jernite, Plu, Xu, Le~Scao, Gugger, Drame, Lhoest, and Rush}]{wolf-etal-2020-transformers}
Thomas Wolf, Lysandre Debut, Victor Sanh, Julien Chaumond, Clement Delangue, Anthony Moi, Pierric Cistac, Tim Rault, Remi Louf, Morgan Funtowicz, Joe Davison, Sam Shleifer, Patrick von Platen, Clara Ma, Yacine Jernite, Julien Plu, Canwen Xu, Teven Le~Scao, Sylvain Gugger, Mariama Drame, Quentin Lhoest, and Alexander Rush. 2020.
\newblock \href {https://doi.org/10.18653/v1/2020.emnlp-demos.6} {Transformers: State-of-the-art natural language processing}.
\newblock In \emph{Proceedings of the 2020 Conference on Empirical Methods in Natural Language Processing: System Demonstrations}, pages 38--45, Online. Association for Computational Linguistics.

\bibitem[{Xiao and Wang(2021)}]{xiao-wang-2021-hallucination}
Yijun Xiao and William~Yang Wang. 2021.
\newblock \href {https://doi.org/10.18653/v1/2021.eacl-main.236} {On hallucination and predictive uncertainty in conditional language generation}.
\newblock In \emph{Proceedings of the 16th Conference of the European Chapter of the Association for Computational Linguistics: Main Volume}, pages 2734--2744, Online. Association for Computational Linguistics.

\bibitem[{Xu et~al.(2023)Xu, Agrawal, Briakou, Martindale, and Carpuat}]{10.1162/tacl_a_00563-hallucination-MT}
Weijia Xu, Sweta Agrawal, Eleftheria Briakou, Marianna~J. Martindale, and Marine Carpuat. 2023.
\newblock \href {https://doi.org/10.1162/tacl_a_00563} {{Understanding and Detecting Hallucinations in Neural Machine Translation via Model Introspection}}.
\newblock \emph{Transactions of the Association for Computational Linguistics}, 11:546--564.

\bibitem[{Yang et~al.(2022)Yang, Gupta, Upadhyay, He, Goel, and Paul}]{yang-etal-2022-tableformer}
Jingfeng Yang, Aditya Gupta, Shyam Upadhyay, Luheng He, Rahul Goel, and Shachi Paul. 2022.
\newblock \href {https://doi.org/10.18653/v1/2022.acl-long.40} {{T}able{F}ormer: Robust transformer modeling for table-text encoding}.
\newblock In \emph{Proceedings of the 60th Annual Meeting of the Association for Computational Linguistics (Volume 1: Long Papers)}, pages 528--537, Dublin, Ireland. Association for Computational Linguistics.

\bibitem[{Zhang et~al.(2023)Zhang, Press, Merrill, Liu, and Smith}]{zhang2023languageModelSnowball}
Muru Zhang, Ofir Press, William Merrill, Alisa Liu, and Noah~A Smith. 2023.
\newblock How language model hallucinations can snowball.
\newblock \emph{arXiv preprint arXiv:2305.13534}.

\bibitem[{Zhang* et~al.(2020)Zhang*, Kishore*, Wu*, Weinberger, and Artzi}]{bert-score}
Tianyi Zhang*, Varsha Kishore*, Felix Wu*, Kilian~Q. Weinberger, and Yoav Artzi. 2020.
\newblock \href {https://openreview.net/forum?id=SkeHuCVFDr} {Bertscore: Evaluating text generation with bert}.
\newblock In \emph{International Conference on Learning Representations}.

\bibitem[{Zhao et~al.(2020)Zhao, Cohen, and Webber}]{zhao2020reducing-quantity-hallucinations}
Zheng Zhao, Shay~B Cohen, and Bonnie Webber. 2020.
\newblock Reducing quantity hallucinations in abstractive summarization.
\newblock In \emph{Findings of the Association for Computational Linguistics: EMNLP 2020}, pages 2237--2249.

\bibitem[{Zhou et~al.(2021)Zhou, Gopalakrishnan, Hedayatnia, Kim, Pujara, Ren, Liu, and Hakkani-Tür}]{Zhou2021-dialogGeneration}
Pei Zhou, Karthik Gopalakrishnan, Behnam Hedayatnia, Seokhwan Kim, Jay Pujara, Xiang Ren, Yang Liu, and Dilek Hakkani-Tür. 2021.
\newblock \href {https://www.amazon.science/publications/commonsense-focused-dialogues-for-response-generation-an-empirical-study} {Commonsense-focused dialogues for response generation: An empirical study}.
\newblock In \emph{SIGDIAL 2021}.

\end{thebibliography}
\bibliographystyle{acl_natbib}

% ========================================

\newpage

\appendix

\section{Appendix}
\label{sec:appendix}

% ------

\subsection{Implementation Details.}
\label{appendix:expt_details}

We implement all of our methods with the \texttt{transformers} package \cite{wolf-etal-2020-transformers}. We mainly follow \citeauthor{ribeiro-etal-2021-investigating-LM-G2T-Generation} to train and test our models in all experiments. In this section we describe all hyperparameters used for reproducibility.

We train BART-large and T5-large as the base generators. They have 406M and 770M parameters, respectively.

\subsubsection{WebNLG} 
\noindent \textbf{BART.} Following \cite{ribeiro-etal-2021-investigating-LM-G2T-Generation}, we add the special tokens \texttt{<H>}, \texttt{<R>}, and \texttt{<T>} to the models’ vocabulary, insert them before the subject, relation, and object, respectively, before concatenating them all into a triple string. We then concatenate all triple strings within an instance to form the input. We train a BART-large model \cite{lewis-etal-2020-bart} as our generator with 2 epochs and a batch size 4. We set the learning rate to be $3\cdot 10^{-5}$. Similar to \cite{ribeiro-etal-2021-investigating-LM-G2T-Generation}, we employ a linearly decreasing learning rate schedule without warm-up. We use beam search as the baseline and set the beam search size to 5. The best checkpoint is selected based on the validation BLEU score \cite{papineni-etal-2002-bleu}. We set the max generation length to 384.

\noindent \textbf{T5.} We perform the same preprocessing as above for T5's input. We additionally append a prefix, ``\textit{translate from Graph to Text:}'' at the beginning of an input. We train a T5-large generator with 10 epochs and batch size 4. We use the same learning rate as suggested in \cite{ribeiro-etal-2021-investigating-LM-G2T-Generation} at $3\cdot 10^{-5}$. We also use a linearly decreasing learning rate schedule without warm-up. Again, the beam search is used as the baseline but the beam size is set to 3. The best checkpoint is again selected based on the validation BLEU score \cite{papineni-etal-2002-bleu}. We set the max generation length also to 384. 

\noindent \textbf{TWEAK Decoding}
For both models, when applying TWEAK decoding, we set the beam size to 4, and generate forward hypotheses using greedy decoding for efficiency. The weighting parameter $\alpha$ is set to 8. We also set the max generation length to 384. 

\subsubsection{TekGen}

We use the same hyperparameters as we do for WebNLG to train and test for both BART-large and T5-large generators. When applying TWEAK decoding, we set $\alpha$ to 8 and 1 for BART and T5, respectively. We use the same beam size as the beam search baseline, where beam size is 5 and 3 for BART-large and T5-large, respectively. Max generation length is also set to 384.

\subsubsection{GenWiki}

We use the same hyperparameters as we do for WebNLG to train and test for both BART-large and T5-large generators, except we set a larger batch size at 32. We also raise the batch size for training T5-large to 16, and keep the other hyperparameters the same as when we train T5-large on WebNLG.

When applying TWEAK decoding, we set $\alpha$ to 8 and 2 for BART-large and T5-large, respectively. We use the same beam size as the beam search baseline, where the beam size is 5 and 3 for BART-large and T5-large, respectively. Max generation length is also set to 384.

% ------

\subsection{Example: Faithful Output is Worded Differently}
\label{appendix:example}

The following shows the output from BART-large using the baseline beam search decoding and the TWEAK-NLI-B+F variant decoding strategy. Although the output of the latter produces a higher faithfulness score (FactKB), it is worded more differently with respect to the reference, resulting in a lower quality score.

\begin{itemize}
    \item \textbf{Facts}: 
    \begin{itemize}
        {\footnotesize
        \item \texttt{(Aston Martin V8, related Mean Of Transportation, Aston Martin DBS)}
        \item \texttt{(Aston Martin V8, engine, 5.3 litres)}
        \item \texttt{(Aston Martin V8, assembly, United Kingdom)}}
    \end{itemize}
    \item \textbf{Reference}: \textit{The Aston Martin V8 is assembled in the United Kingdom and has an engine volume of 5.3 litres. The Aston Martin V8 and Aston Martin DBS are a related means of transport.}
    \item \textbf{Beam search}: \textit{Aston Martin V8, which is 5.3 litres and made in the United Kingdom, is related to the Aston Martin DBS.}
    \item \textbf{TWEAK-NLI-B+F}: \textit{The United Kingdom is the location of the assembly of the Aston Martin V8 which has a 5.3 litres engine and is related to the DBS.}
\end{itemize}

\section{FATE Dataset Statistics}
\label{appendix:fate-stats}
% Please add the following required packages to your document preamble:
% \usepackage{multirow}
\begin{table}[H]
\resizebox{\linewidth}{!}{
\begin{tabular}{lrrrrrr}
\toprule
\multicolumn{1}{c}{\multirow{2}{*}{\textbf{FATE}}} & \multicolumn{1}{c}{\multirow{2}{*}{\textbf{Subj}}} & \multicolumn{1}{c}{\multirow{2}{*}{\textbf{Rel}}} & \multicolumn{1}{c}{\multirow{2}{*}{\textbf{Obj}}} & \multicolumn{1}{c}{\multirow{2}{*}{\textbf{Triples}}} & \multicolumn{2}{c}{\textbf{Entity Avg.}} \\ \cmidrule{6-7} 
\multicolumn{1}{c}{}                               & \multicolumn{1}{c}{}                               & \multicolumn{1}{c}{}                              & \multicolumn{1}{c}{}                              & \multicolumn{1}{c}{}                                  & \textbf{Triples}     & \textbf{Words}    \\ \midrule
Original                                           & 423                                                & 235                                               & 1499                                              & 922                                                   & 4.54                 & 19.8              \\
Perturbed                                     & 432                                                & 1666                                              & 3118                                              & 7368                                                  & 17.05                & 20.0                \\ \bottomrule
\end{tabular}
}
\caption{Dataset statistics for our curated FATE. Both the original and the perturbed sets contain 18,102 instances. All numbers are counts of unique instances.}
\label{tab:fate}
\end{table}

\section{Statistics of Evaluation Benchmarks}
\label{appendix:eval-dataset-stats}
% Please add the following required packages to your document preamble:
% \usepackage{multirow}
\begin{table}[h!]
\resizebox{\linewidth}{!}{
\begin{tabular}{cr|rrrrrr}
\toprule
\multicolumn{2}{c}{\multirow{2}{*}{\textbf{Dataset}}} & \multicolumn{1}{c}{\multirow{2}{*}{\textbf{Subj}}} & \multicolumn{1}{c}{\multirow{2}{*}{\textbf{Rel}}} & \multicolumn{1}{c}{\multirow{2}{*}{\textbf{Obj}}} & \multicolumn{1}{c}{\multirow{2}{*}{\textbf{Triples}}} & \multicolumn{2}{c}{\textbf{Entity Avg.}} \\ \cmidrule{7-8} 
\multicolumn{2}{c}{}                                  & \multicolumn{1}{c}{}                               & \multicolumn{1}{c}{}                              & \multicolumn{1}{c}{}                              & \multicolumn{1}{c}{}                                  & \textbf{Triples}     & \textbf{Words}    \\ \midrule
\multirow{2}{*}{\textbf{WebNLG}}             & Train           & 430                                                & 246                                               & 1613                                              & 2090                                                  & 4.8                  & 19.8              \\
                                    & Test            & 575                                                & 300                                               & 1882                                              & 2331                                                  & 4.0                    & 19.5              \\ \midrule
\multirow{2}{*}{\textbf{TekGen}}             & Train           & 20K                                                & 1K                                                & 13K                                               & 34K                                                   & 1.7                  & 21.0                \\
                                    & Test            & 1000                                               & 200                                               & 1176                                              & 1783                                                  & 1.7                  & 21.4              \\ \midrule
\multirow{2}{*}{\textbf{GenWiki}}            & Train           & 713K                                               & 287                                               & 273K                                              & 1754K                                                 & 2.4                  & 29.2              \\
                                    & Test            & 817                                                & 157                                               & 2150                                              & 1783                                                  & 3.9                  & 18.6              \\ \bottomrule
\end{tabular}
}
\caption{Dataset statistics for WebNLG, TekGen, and GenWiki. All numbers are counts of unique instances.}
\label{tab:dataset_stats}
\end{table}

% \section{Out-of-distribution Evaluation for T5}
% We show the out-of-distribution evaluation for T5 on TekGen and GenWiki in Table~\ref{tab:ood_results_T5}.

% \label{appdix:ood-result-t5}
% \begin{table*}[htb!]
% \centering
% \resizebox{0.9\linewidth}{!}{
% \begin{tabular}{cl|c|ccc|c|ccc}
% \toprule

% \end{tabular}
% }
% \caption{Generalization results on \textit{out-of-distribution} (OOD) test sets TekGen and GenWiki for T5-large. \textbf{BS} = BERTScore. Numbers in \textbf{bold} are the highest scores among the baselines (greedy and beam) or among the TWEAK variants.}
% \label{tab:ood_results_T5}
% \end{table*}

\section{Weighting Effects in Out-of-distribution Evaluations}
\label{appendix:weighting-effect-ood}

\begin{figure*}
    \centering
    \includegraphics[width=\linewidth]{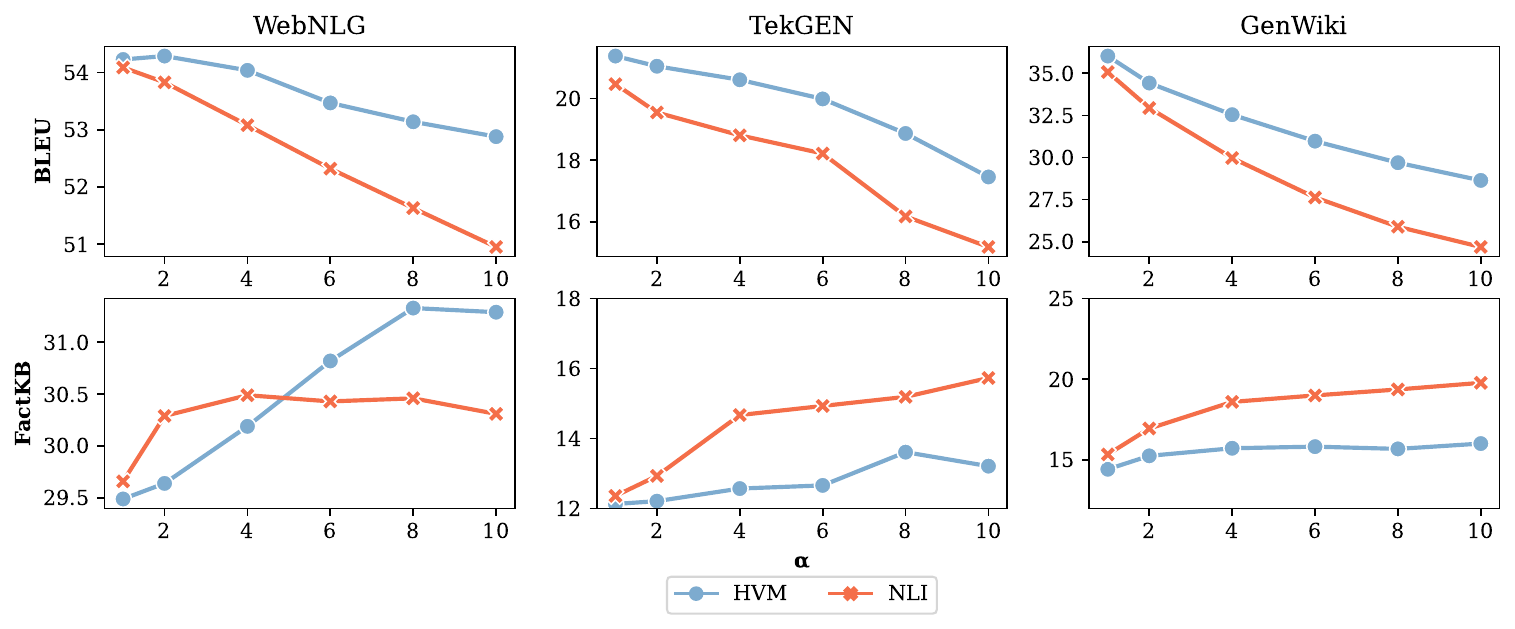}
    \caption{The effect on quality (BLEU) and faithfulness (FactKB) from choosing different $\alpha$ in Equ.~\eqref{equ:tweak_scoring}, with $\alpha = 0$ being equivalent to beam search. The results are obtained using TWEAK-NLI-B+F and TWEAK-HVM variants on WebNLG, TekGen and GenWiki with BART.}
    \label{fig:weighting-effect-ood}
\end{figure*}

We have discussed the effect of manipulating weighting coefficient $\alpha$ in in-distribution experiments in Sec.~\ref{sec:analysis}. We further plot the weighting effect on \textit{out-of-distribution} (OOD) datasets in Figure~\ref{fig:weighting-effect-ood}. On the two other datasets, HVM underperforms NLI on faithfulness due to distribution shift, but maintains higher quality scores than NLI at all $\alpha$ values. This shows HVM maintains the quality edge over NLI even in the OOD settings.

\section{Prompt Template to Generate FATE}
\label{appendix:prompt-template}

In this section we show the prompt templates we use with the LLM to create our FATE dataset (Sec.~\ref{subsec:fate}): one prompt is used to perturb a fact triple (Figure~\ref{gpt-prompt-edit-triple}), and the other is to generate the description for the perturbed fact triple (Figure~\ref{gpt-prompt-edit-sentence}). Different from the majority of the similar methods deployed in the literature focusing solely on subjects or objects, we allow perturbations to happen at all possible positions -- subject, relation, and object -- in order to obtain more diverse datapoints.

\section{More Examples}
\label{appendix:more-qualitative-examples}

We show three more examples in Table~\ref{tab:qualitative-examples-table}:
\begin{itemize}
    \item \textbf{First example:} All three decoding strategies produce faithful textual descriptions for the given triples. TWEAK-HVM's output is arguably more readable than the others, but in terms of textual similarity with the reference, both TWEAK variants produce less similar output than beam search.
    \item \textbf{Second example:} Beam search generates an output asserting \textit{"Costa Crociere is the parent company of Carnival Corporation"}, which directly contradicts the input fact 2 (\textit{"The parent company of Costa Crociere is Carnival Corporation"}). Both TWEAK variants avoid this mistake.
    \item \textbf{Third example:} Beam search misses the fact "\textit{nearest city}" which is captured by our TWEAK variants.
\end{itemize}

\begin{table*}[t!]
\centering
\begin{tabular}{p{0.98\linewidth}}
   \toprule 
  \textbf{Fact triples:} 
  \begin{enumerate}
      \setlength{\itemsep}{-0.5em}
      \item <H> Accademia di Architettura di Mendrisio <R> country <T> Switzerland
      \item <H> Accademia di Architettura di Mendrisio <R> dean <T> Mario Botta
      \item <H> Accademia di Architettura di Mendrisio <R> city <T> Mendrisio
      \item <H> Accademia di Architettura di Mendrisio <R> established <T> 1996
      \item <H> Accademia di Architettura di Mendrisio <R> academic Staff Size <T> 100
      \item <H> Accademia di Architettura di Mendrisio <R> location <T> Ticino
  \end{enumerate}
  \\
  \midrule
  \textbf{Reference:} Accademia di Architettura di Mendrisio in Mendrisio, Switzerland has 100 employees, was established in 1996 and is overseen by Dean Mario Botta. The school's location is Ticino. \\ 

  \textbf{Beam Search:} The Accademia di Architettura di Mendrisio is located in Ticino, Mendrisio, Switzerland. It was established in 1996 and its dean is Mario Botta. It has 100 academic staff. \\
  
  \textbf{TWEAK-NLI:} Mario Botta is Dean of the Accademia di Architettura di Mendrisio in Mendrisio, Ticino, Switzerland. The Accademia was established in 1996 and has 100 academic staff. \\
  
  \textbf{TWEAK-HVM:} The dean of the Accademia di Architettura di Mendrisio is Mario Botta. The Accademia was established in 1996 in the city of Mendrisio, Ticino, Switzerland and has 100 academic staff. \\ \midrule \midrule
  \textbf{Fact triples:} 
  \begin{enumerate}
      \setlength{\itemsep}{-0.5em}
      \item <H> Costa Crociere <R> location <T> Genoa 
      \item <H> Costa Crociere <R> parent Company <T> Carnival Corporation \& plc 
      \item <H> AIDAstella <R> operator <T> AIDA Cruises 
      \item <H> AIDAstella <R> builder <T> Meyer Werft
      \item <H> AIDAstella <R> owner <T> Costa Crociere
  \end{enumerate}
  \\
  \midrule
  \textbf{Reference:} Carnival Corporation \& plc is the parent company of Costa Crociere in Genoa, who own the AIDAstella. AIDAstella was built by Meyer Werft and is operated by AIDA Cruises. \\ 

  \textbf{Beam Search:} AIDAstella, built by Meyer Werft and owned by Costa Crociere, is operated by AIDA Cruises. Costa Crociere is the parent company of Carnival Corporation \& plc and is located in Genoa. \\
  
  \textbf{TWEAK-NLI:} AIDAstella, built by Meyer Werft, is operated by AIDA Cruises. It is owned by Costa Crociere, whose parent company is Carnival Corporation \& plc and is located in Genoa. \\
  
  \textbf{TWEAK-HVM:} The parent company of Costa Crociere, located in Genoa, is Carnival Corporation \& plc. AIDAstella, operated by AIDA Cruises, was built by Meyer Werft and is owned by Costa Crociere. \\ \midrule \midrule
  \textbf{Fact triples:} 
    \begin{enumerate}
      \setlength{\itemsep}{-0.5em}
      \item <H> 14th New Jersey Volunteer Infantry Monument <R> district <T> Monocacy National Battlefield
      \item <H> Monocacy National Battlefield <R> nearest City <T> Frederick Maryland
  \end{enumerate}
  \\
  \midrule
  \textbf{Reference:} The 14th New Jersey Volunteer Infantry Monument is near Frederick, Maryland in the district of the Monocacy National Battlefield.	 \\ 

  \textbf{Beam Search:} The 14th New Jersey Volunteer Infantry Monument is located in the district of the Monocacy National Battlefield, Frederick, Maryland. \\
  
  \textbf{TWEAK-NLI:} Frederick, Maryland is the nearest city to Monocacy National Battlefield where the 14th New Jersey Volunteer Infantry Monument is located. \\
  
  \textbf{TWEAK-HVM:} The 14th New Jersey Volunteer Infantry Monument is in the district of the Monocacy National Battlefield and the nearest city is Frederick, Maryland. \\ 
  
  \bottomrule
\end{tabular}
\caption{Three examples taken from T5-large's outputs on WebNLG. }
\label{tab:qualitative-examples-table}
\end{table*}

\onecolumn
\begin{tcolorbox}[tab2,tabularx={X},title=Prompt template for perturbing original facts.,boxrule=0.5pt]
Using your commonsense knowledge to edit the predicate in the old triple to make it counterfactual. Note that you should not always use predicate negation.

\quad

Old triple: ('Aarhus Airport', 'operating Organisation', 'Aarhus Lufthavn A/S')

New triple: ('Aarhus Airport', 'leader Name', 'Aarhus Lufthavn A/S')

\quad

Old triple: ('Aarhus Airport', 'location', 'Tirstrup')

New triple: ('Aarhus Airport', 'country', 'Tirstrup')

\quad

Old triple: ('Aarhus Airport', 'location', 'Tirstrup')

New triple: ('Aarhus Airport', 'birthday', 'Tirstrup')

\quad

Old triple: ("jamaica at the fifa world cup", "subclass of", "jamaica national football team")

New triple: ("jamaica at the fifa world cup", "president of", "jamaica national football team")

\quad

Old triple: ("kentucky louisville rivalry", "participating team", "louisville cardinals")

New triple: ("kentucky louisville rivalry", "beat", "louisville cardinals")

\quad

Old triple: \{\$old\_triple\}

New triple:
\end{tcolorbox}
\noindent\begin{minipage}{\textwidth}
\captionof{figure}{Prompt template we used for perturbing a fact triple. We allow the model to perturb both subject/object and predicate.}\label{gpt-prompt-edit-triple}
\end{minipage}

\begin{tcolorbox}[tab2,tabularx={X},title=Prompt template for editing description to align with the new fact.,boxrule=0.5pt]
Minimally edit the following sentence so it supports the new fact triple instead of the old fact triple, while highlighting your edited text spans with '[' and ']'.

\quad

Sentence: Aarhus Airport serves the city of Aarhus, Denmark.

Old fact: ('Aarhus Airport', 'city Served', 'Aarhus Denmark')

New fact: ('Taylor County Texas', 'city Served', 'Aarhus Denmark')

Revised: [Taylor County Texas] swerves the city of Aarhus, Denmark

\quad

Sentence: Aarhus Airport is operated by Aarhus Lufthavn A/S.

Old fact: ('Aarhus Airport', 'operating Organisation', 'Aarhus Lufthavn A/S')

New fact: ('Aarhus Airport', 'death Date', 'Aarhus Lufthavn A/S')

Revised: Aarhus Airport\'s [death date is] Aarhus Lufthavn A/S

\quad

Sentence: The location of Aarhus Airport is Tirstrup.

Old fact: ('Aarhus Airport', 'location', 'Tirstrup')

New fact: ('Aarhus Airport', 'leader Name', 'Tirstrup')

Revised: The [leader name of] Aarhus Airport is Tirstrup.

\quad

Sentence: \{\$sentence\}

Old fact: \{\$old\_triple\}

New fact: \{\$new\_triple\}

Revised: 
\end{tcolorbox}
\noindent\begin{minipage}{\textwidth}
\captionof{figure}{Prompt template we used for editing textual description to align with the edited fact, while annotating the edited span.}\label{gpt-prompt-edit-sentence}
\end{minipage}

\twocolumn

\end{document}